%% file: main.tex
\documentclass[sigconf,nonacm]{acmart}

\acmConference[PVLDB Vol.~20]{Proceedings of the VLDB Endowment, Volume 20}{2027}{}
\acmYear{2027}

\usepackage{booktabs}
\usepackage{xcolor}
\usepackage{enumitem}
\usepackage{multirow}
\usepackage{graphicx}
\usepackage{subcaption}
\usepackage{amsmath}

\usepackage{amssymb}
\usepackage{amsthm}
\usepackage{algorithm}
\usepackage{algpseudocode}
\usepackage{xcolor}
\usepackage{hyperref}

\newtheorem{definition}{Definition}

\newcommand{\sys}{\textsc{L2C-TFM}}
\newcommand{\tabpfn}{TabPFN~v2}
\newcommand{\Pdirty}{P_{\mathrm{dirty}}}
\newcommand{\Psynth}{P_{\mathrm{synth}}}
\newcommand{\Mismatch}{\mathcal{M}}
\newcommand{\ours}{\emph{TFMAwareReward}}

\newif\ifrevtracking \revtrackingfalse
\newcommand{\rev}[1]{\ifrevtracking\textcolor{blue}{#1}\else#1\fi}
\newcommand{\rk}[1]{\ifrevtracking\textcolor{blue}{#1}\else#1\fi}
\newcommand{\rg}[1]{\ifrevtracking\textcolor{blue}{#1}\else#1\fi}

\begin{document}

\title{Model-Aware Data Cleaning for Tabular Foundation Models}

\author{Laure Berti-Equille}
\affiliation{\institution{IRD, ESPACE-DEV}
  \streetaddress{500 rue Jean-François Breton}
  \city{Montpellier}
  \postcode{34093}
  \country{France}
}
\email{laure.berti@ird.fr}

\begin{abstract}
Tabular Foundation Models~(TFMs) achieve state-of-the-art zero-shot accuracy on small
tabular datasets, but their in-context learning assumes approximately clean inputs:
real-world missing values, outliers, and duplicates create a \emph{prior mismatch} that
degrades both accuracy and calibration. We study reinforcement learning for tabular data cleaning---a learned policy that
sequences cleaning operators---and introduce \sys{} with a \emph{model-aware} reward
(\ours{}). We are explicit about what this reward optimizes: it regularizes the
Wasserstein distance between the cleaned and the original (dirty) data---a
distributional-stability term---\emph{not} the distance to the TFM's synthetic training
prior, which we measure only as a diagnostic. Across six experiments on ten OpenML
datasets: (i)~three of seven reward designs collapse to degenerate strategies, so reward
engineering is non-trivial; (ii)~under an 8-seed repeated-holdout protocol the
model-aware reward \emph{matches} a random-forest-reward baseline on accuracy
($p{=}0.38$), with a benefit confined to minority-class macro-$F_1$ under class imbalance
that is partly a reward-agnostic calibrated-threshold effect; and (iii)~a policy
pre-trained on one dataset transfers to held-out datasets. A diagnostic analysis shows that the two
distances are distinct objectives: cleaning tends to move data \emph{away} from the prior,
and prior-distance---not distance-to-dirty---is what tracks downstream quality. We therefore
treat prior alignment as a motivating objective and a target for future work, not a property
of the reward evaluated here.
Code, datasets, and the nested evaluation harness are available at
\url{https://github.com/LaureBerti/Learn2Clean/tree/master/Learn2Clean_TFM}.
\end{abstract}



\maketitle

\section{Introduction}
\label{sec:intro}

A practitioner who deploys Tabular Foundation Models~(TFMs) such as \tabpfn{}~\cite{hollmann2025tabpfn} and
TabICL~\cite{qu2025tabicl}  on a medical
dataset with 15\% missing values does not face only a data quality problem ---
they face a \emph{prior mismatch problem}.
\tabpfn{} achieves zero-shot accuracy by simulating synthetic data
generation processes at meta-training time; its internal prior $\Psynth$ assumes
approximately clean Gaussian-marginal inputs with low missing-value rates.
The dirty empirical distribution $\Pdirty$ violates these assumptions in every
corrupt column, degrading both predictive accuracy and confidence calibration.
Standard remedies such as mean imputation and min-max scaling reduce surface
noise but introduce their own distributional distortions; neither account for
the z-normalization and power-law scaling \tabpfn{} applies at inference time.

\textbf{The prior mismatch problem.}
TFMs differ from classical learners in a critical way:
their prior $\Psynth$ encodes implicit statistical assumptions about the
structure of their inputs.
When a dirty dataset --- with missing values, outliers, duplicates, or
distributional shift --- is fed to a TFM, the gap
$\Mismatch(D) = d(\Pdirty(D),\,\Psynth)$ penalizes prediction quality and
inflates calibration error simultaneously.
The \emph{direction} of mismatch matters as much as its magnitude.
Consider removing $30\%$ rows to eliminate outliers: the distributional
gain may be marginal if most outliers are mild, but the context of in-context learning
(ICL) shrinks by 30 entries -- disproportionately costly because the
predictive uncertainty (standard deviation) of ICL scales as $\mathcal{O}(1/\sqrt{n})$
with context size $n$.
This follows from the Bayesian interpretation of ICL~\cite{xie2022icl}: the
implicit posterior over the latent concept concentrates as examples accumulate;
by the Bernstein--von Mises theorem \cite{vandervaart1998asymptotic}, its standard deviation contracts as
$\mathcal{O}(1/\sqrt{n})$, so each deleted row inflates predictive uncertainty
non-linearly rather than proportionally.
\tabpfn{} exemplifies this regime, operating on datasets of tens to hundreds
of rows where the curvature of the $1/\sqrt{n}$ curve is steepest~\cite{hollmann2025tabpfn}.
No existing cleaning framework reasons about this row-count penalty during
pipeline construction.

\textbf{Why automate data cleaning and why reinforcement learning.}
Data cleaning consumes an estimated 60--80\% of a data scientist's project
time~\cite{krishnan2016activeclean}, yet most tooling remains manual
(OpenRefine, expert transformation rules) or relies on static, fixed-order
preprocessing pipelines that apply the same operations regardless of the
specific error profile.
Rule-based systems such as constraint-driven repair~\cite{rekatsinas2017holoclean}
and ensemble error detectors~\cite{mahdavi2019raha} perform well when errors
are governed by explicit integrity constraints, but fail on the open-ended,
dataset-specific error combinations typical of real-world tabular data.
Reinforcement learning is a principled match for cleaning because cleaning is
an inherently \emph{sequential} decision problem: imputing missing values before
or after outlier removal changes the data distribution seen by every subsequent
operator, and the optimal ordering depends on the specific error profile in ways
no static rule can anticipate.
A trained RL policy amortises the cost of exploring this combinatorially large
operator-sequencing space into a reusable artifact that generalises across
datasets without retraining --- as our transfer learning experiments
demonstrate~(Section~\ref{sec:exp_c6}).
The difficulty, however, lies not in the learning algorithm itself but in the
\emph{reward signal}: unlike supervised learning where the correct output is
known, cleaning has no ground-truth clean data to supervise against.
The reward must reward distributional alignment with the TFM's prior, penalize
degenerate strategies (row deletion, trivial imputation), and remain tractable
for online evaluation within RL episodes.
Designing such a reward --- and understanding which designs fail and why ---
is the central scientific problem this paper solves.

\textbf{Why reward design is harder than it looks.}
A natural approach is to evaluate each candidate cleaning pipeline by running
\tabpfn{} on the clean data and using the output accuracy as a reward.
Yet simple accuracy rewards fail in practice: a pipeline that deletes every
row with a missing value receives perfect completeness on the remaining
data --- but may score high solely because it retained only the easy
examples.
Our greedy-oracle experiments on 10 datasets show that \textbf{three of seven}
candidate reward functions collapse to such degenerate strategies, while
another produces ceiling-valued rewards that cannot distinguish between
pipelines.
Reward design for TFM-aligned cleaning is a scientific problem in its own right,
not an engineering detail.

\textbf{Why RL, not exhaustive search.}
Even a modest pipeline space is expensive to search: our extended action
space yields up to 834 valid operator sequences of up to three steps per
dataset, and our reward-taxonomy analysis exhaustively evaluates 112 sequences
across 10 datasets and 7 reward functions --- already requiring thousands
of evaluations.
Cleaning is a \emph{sequential decision problem}: imputing
before or after outlier removal, and with which sub-parameters, changes the
distribution the next operator sees.
Greedy search has no credit-assignment mechanism across steps and cannot
generalise to unseen datasets.
A trained RL policy amortises the search cost and --- as our transfer learning
experiments show~(Section~\ref{sec:exp_c6}) --- carries \rev{the learned model-aware cleaning policy} to new data domains
without retraining from scratch.

We introduce \rev{\textbf{\sys{}} (Learn2Clean for TFM)}, a deep RL framework for
\emph{model-aware} tabular data cleaning for TFMs, motivated by the goal of prior alignment.
Building on Learn2Clean V1~\cite{berti2019learn2clean} and the broader
machine-learning-to-data-management research agenda~\cite{bertieq2018mlroundtrip},
\sys{} replaces tabular Q-learning with deep policy networks (PPO/DQN/A2C via
Stable-Baselines3~\cite{raffin2021stable}), a structured data-quality observer
and profiler that provides a 9-dimensional state vector capturing Wasserstein
drift, skewness, kurtosis, class balance, and action history, a
parameterized action space with typed sub-parameters, and a novel \ours{}
reward evaluated directly against \tabpfn{} with a quadratic context-size
penalty.
\rg{As a diagnostic---not part of the implemented reward---we further introduce a
prior-referenced distance to an SCM approximation of \tabpfn{}'s
prior; \emph{smaller} distance to this prior is associated with \emph{higher} downstream macro-$F_1$,
and it accounts for the association between the implemented drift-from-dirty term and $F_1$
(Section~\ref{sec:priordiag}).}
The framework is evaluated on ten OpenML benchmark datasets~\cite{bischl2021openml}
across six experiments, each targeting a distinct design question.
Readers unfamiliar with RL may treat each of the six experiments as a controlled
ablation of a distinct cleaning-policy design choice, constituting our contributions:
\textbf{(C1)}~a taxonomy of seven reward designs, three of which collapse to degenerate
strategies---principled reward engineering is non-trivial; \textbf{(C2)}~a comparison of
TFM- vs.\ RF-aligned rewards that \rev{reshapes pipeline selection on all ten datasets (the two
rewards agree on only $31\%$ of dataset--seed cells)}---\rev{under a
holdout protocol it \emph{matches} the RF-reward on accuracy, with a macro-$F_1$ benefit
under class imbalance that is partly a reward-agnostic threshold effect}; \textbf{(C3)}~calibration (ECE) recovery across four error types;
\textbf{(C4)}~an error-rate sensitivity study; \textbf{(C5)}~parameterized cleaning actions;
and \textbf{(C6)}~cross-dataset transfer of a policy pre-trained on a single source dataset.
All experiments use ten OpenML benchmark datasets with synthetically injected errors \rev{over eight evaluation seeds}. \rev{We use controlled synthetic corruption deliberately: it is the setting that provides reliable ground truth for isolating the effect of each error type, which naturally-dirty data cannot. To probe external validity we additionally include three naturally-dirty datasets (\texttt{EEG Eye-State}, \texttt{Titanic}, \texttt{Animal Shelter}), whose errors are intrinsic rather than injected.}

The remainder of the paper is organised as follows.
Section~\ref{sec:related} surveys related work on RL-\rev{/ non-RL}-based data cleaning,
tabular foundation models, and calibration.
Section~\ref{sec:problem} formalises prior mismatch and the cleaning \rev{Markov Decision Process (MDP)}.
Section~\ref{sec:framework} describes the \sys{} framework, observer, reward
suite, and parameterized action space.
Section~\ref{sec:experiments} presents all six experiments.
Section~\ref{sec:discussion} discusses scope and limitations, and
Section~\ref{sec:conclusion} concludes.

\section{Related Work}
\label{sec:related}

\textbf{Data Pipeline Optimisation.}
Automated machine learning (AutoML) frames pipeline construction as a
combinatorial search problem, using Bayesian optimisation to select and chain
preprocessing and modelling steps~\cite{feurer2015autosklearn} or genetic
programming to evolve full pipelines~\cite{olson2016tpot,drori2021alphad3m}.
These systems treat the data as a fixed input and optimise over model and
hyperparameter choices; they do not reason about cleaning as a first-class
sequential decision.
RL is a natural fit for \emph{sequential} pipeline decisions: each cleaning
operator changes the data distribution seen by subsequent steps, and
an \rev{MDP} formulation makes this sequential dependency
explicit while providing credit assignment across steps --- something greedy
search and Bayesian optimisation cannot provide.
\sys{} inherits the RL-for-pipelines framing but narrows the operator set to
data-cleaning actions and evaluates pipeline quality through a TFM, making the
reward itself model-aware and prior-distribution-sensitive.
The parameterized action space~(C5) goes beyond operator selection to
typed sub-parameter optimisation (discrete operator counts and a continuous
outlier threshold), a dimension absent from prior RL-for-pipeline work.

\textbf{RL for Data Cleaning.} Cost-utility frameworks first prioritised cleaning
operations~\cite{bertieq2007dqaware}, which \sys{} recasts as a \emph{learnable} reward.
Learn2Clean~V1~\cite{berti2019learn2clean} pioneered RL cleaning via tabular Q-learning; \sys{}
replaces it with deep policies, seven rewards, and a TFM (rather than fixed) evaluator.
ActiveClean~\cite{krishnan2016activeclean} introduced model-aware iterative cleaning; \sys{} targets a
TFM, adds drift regularisation, and learns a transferable policy. Other systems optimise a \emph{fixed}
downstream model: RLclean~\cite{wu2024rlclean} (multi-table, graph state), ReClean~\cite{kim2024reclean}
(contextual-bandit detection), RAHA/BARAN~\cite{mahdavi2019raha} (detect-then-repair),
HoloClean~\cite{rekatsinas2017holoclean} (constraint-based probabilistic repair), and
CleanSurvival~\cite{zhao2025cleansurvival} (survival-analysis reward shaping); none address
in-context-learning context-size effects. \sys{} differs by targeting TFM accuracy \emph{and} ECE
jointly (C2, C3), showing the reward reshapes pipeline selection without improving accuracy (C2), and
demonstrating cross-dataset transfer of \rev{the model-aware cleaning policy} (C6).

\rev{\textbf{Non-RL automated data preparation.} A parallel line of work searches
preprocessing pipelines without RL. DiffPrep~\cite{li2023diffprep} relaxes discrete
pipeline search into a differentiable problem optimised by gradient descent for a
fixed differentiable model; SAGA and its journal extension SAGA++~\cite{siddiqi2026saga}
perform scalable evolutionary top-$k$ search over cleaning pipelines for a downstream
model; CleanML~\cite{cleanml2019} benchmarks the effect of individual cleaning operators.
All optimise a \emph{fixed} downstream model and do not reason about a foundation model's
synthetic prior or its $\mathcal{O}(1/\sqrt{n})$ context-size sensitivity. We compare
against them directly in our experiments: on held-out splits, reimplementations
of SAGA-style search and DiffPrep's differentiable search yield no material
accuracy gain over no-cleaning.}

\textbf{Data Quality Profiling and TFM Alignment.}
Classical data profiling tools such as OpenRefine compute per-column statistics
--- missing rates, duplicate fingerprints, value distributions --- to guide
manual cleaning decisions.
The broader literature on exploratory data analysis for data-centric AI
systems~\cite{patel2022eda} and automated anomaly detection in
complex tabular  data~\cite{alnegheimish2022sintel} confirms that
profiling is the critical prerequisite before any cleaning intervention.
\sys{} integrates a lightweight \texttt{DataProfiler} that computes these
signals automatically before each cleaning episode and exposes them as part
of the RL observation vector, allowing the policy to select and mask actions
based on the detected error profile.
This bridges rule-based profiling and learned cleaning: the profiler detects
\emph{what} is wrong; the policy decides \emph{how} to fix it given the TFM's
prior-alignment objective.

The centrality of data quality extends beyond inference to the training regimes
of foundation models: careful filtering and deduplication of pretraining corpora
improve downstream performance independently of
scale~\cite{longpre2023pretrainer,gadre2023datacomp}.
Real-world data quality is multi-dimensional --- no
single cleaning strategy dominates across completeness, consistency, and accuracy
dimensions simultaneously, a finding that directly motivates \sys{}'s
multi-objective reward design.
The emerging data-centric AI paradigm~\cite{zha2023dcai,patel2022eda} frames data quality
improvement --- rather than model architecture search --- as the primary lever
for performance gains.
Crucially, however, none of this prior work addresses how dirty \emph{inference-time}
inputs affect in-context learning in tabular FMs.
\sys{} fills this gap with controlled sensitivity analyses over four error types
(C3) and an MCAR-rate sweep (C4), each on five representative datasets, showing that
dirty inference-time inputs degrade TFM accuracy and calibration and that \rev{model-aware}
cleaning recovers part of the loss---with the effect shaped by both error type and
injection rate.

{\bf{TFM Calibration.}}
\tabpfn{}~\cite{hollmann2025tabpfn} achieves strong zero-shot performance on
small tabular datasets by meta-learning over millions of synthetic
data-generating processes; its confidence calibration is sensitive to
distributional mismatch between its synthetic prior and real inputs.
TabICL~\cite{qu2025tabicl} scales in-context learning to larger tables via
efficient attention mechanisms but similarly degrades when inputs deviate from
the pretraining distribution; \sys{}'s prior-alignment objective applies in
principle to any TFM, with TabICL as a natural extension target once reward
weights are recalibrated for its pretraining prior.
Work on why tree-based models outperform deep networks on irregular tabular
distributions~\cite{grinsztajn2022tree,ye2024closer} underscores the
centrality of input-distribution alignment: \sys{} operationalises this
insight as a learnable cleaning objective rather than a post-hoc observation.
Deep tabular architectures~\cite{gorishniy2021fttransformer} and ensemble
AutoML systems~\cite{erickson2020autogluon} serve as performance reference
points in our evaluation; they are not cleaning-aware.
\sys{} is, to our knowledge, the first system to use a tabular in-context
learning model's forward-pass accuracy \emph{and} calibration jointly as
the RL reward signal for cleaning pipeline search --- extending the
model-aware cleaning paradigm of ActiveClean~\cite{krishnan2016activeclean}
to TFMs and adding explicit distributional drift regularisation.

This distributional sensitivity of TFMs has a direct implication for calibration.
Calibration degrades under distribution shift~\cite{ovadia2019can}, and modern
architectures that appear well-calibrated in-distribution can be overconfident
on shifted inputs~\cite{minderer2021revisiting}.
Post-hoc recalibration techniques such as temperature scaling~\cite{guo2017calibration}
address model-level miscalibration after training but leave input-level corruption
--- the proximate cause of TFM mismatch --- entirely unaddressed.
In our experiments \sys{} leaves \tabpfn{} calibration essentially unchanged (C3): under
the 8-seed nested protocol the reward's ECE change is null ($-0.0003$, $p{=}0.83$) and
no-cleaning is itself a strong-ECE baseline, with only small, error-type-dependent
single-seed differences. Input-level cleaning is thus at best a complementary---not an
established---tool for uncertainty management alongside post-hoc and architectural
calibration methods.

\section{Problem Formulation}
\label{sec:problem}

\subsection{Prior Mismatch}

Let $\mathcal{F}_\theta$ be a TFM parameterized by $\theta$, trained by
meta-learning on datasets sampled from a synthetic prior $\Psynth$.
Let $D = (X, y)$ be a dirty tabular dataset with empirical feature
distribution $\Pdirty(X)$.

\begin{definition}[Prior mismatch]
\label{def:mismatch}
The prior mismatch of dataset $D$ with respect to TFM $\mathcal{F}_\theta$ is
\[
  \Mismatch(D) = d\!\left(\Pdirty(X),\; \Psynth\right),
\]
where $d$ is a distributional divergence.
We instantiate $d$ as the mean column-wise Wasserstein-1 distance normalized
by the reference column standard deviation:
$d(P, Q) = \frac{1}{p}\sum_{j=1}^{p} W_1(P_j, Q_j) / \sigma_j^{\mathrm{ref}}$,
providing a scale-free, bounded measure of marginal distribution shift.
\end{definition}

\subsection{Cleaning as Prior Alignment}

Let $\Pi$ be a set of parameterized cleaning pipelines — ordered sequences of
at most $T$ deterministic actions $a_1, \ldots, a_T$, each with typed
sub-parameters.
Each pipeline $\pi \in \Pi$ maps a dirty dataset $D = (X, y)$ to a clean
version $\pi(D) = (X', y')$.

\begin{definition}[Prior-aligned cleaning]
\label{def:alignment}
The optimal prior-aligned pipeline is
\[
  \pi^* = \arg\min_{\pi \in \Pi}\; \Mismatch(\pi(D))
             \;\text{ s.t. }\; \mathrm{Acc}(\mathcal{F}_\theta, \pi(D)) \geq \tau,
\]
where $\tau$ is a minimum acceptable performance threshold and $\mathrm{Acc}$ is
downstream TFM accuracy on a held-out split. \rev{Explicitly,}
\begin{equation*}
\rev{\mathrm{Acc}(\mathcal{F}_\theta,\pi(D)) =
\frac{1}{|D_{\mathrm{te}}|}\sum_{(x,y)\in D_{\mathrm{te}}}\mathbf{1}[\mathcal{F}_\theta(x){=}y],}
\end{equation*}
\rev{the fraction of held-out rows the fitted TFM $\mathcal{F}_\theta$ predicts correctly
on a stratified 20\% split $D_{\mathrm{te}}$ (class proportions matched to the input),
averaged over 8 seeds.}
\end{definition}

Solving this constrained problem exactly is intractable, and it also requires the
prior $\Psynth$, which is not available in closed form at cleaning time.
The reward we actually implement (Definition~\ref{def:tfm_reward}, Eq.~(1)) therefore
does \emph{not} minimize $\Mismatch$ directly: it combines downstream TFM accuracy
with a \emph{distributional-stability} term that measures distance to the dirty
input $\Pdirty$, not distance to $\Psynth$.
Definition~\ref{def:alignment} is thus the motivating objective we would like to
optimize; whether our accuracy-plus-stability proxy in fact moves data toward
$\Psynth$ is an empirical question, which we examine diagnostically in
Section~\ref{sec:priordiag} and find it does \emph{not} reliably do.
We search over $\Pi$ with deep RL.

\subsection{MDP Formulation}

We model data cleaning as a finite-horizon, episodic
Markov Decision Process $\mathcal{M} = (\mathcal{S}, \mathcal{A}, P, R, \gamma, T)$:

\noindent$\bullet$ \textbf{State} $s_t \in \mathcal{S} \subseteq \mathbb{R}^9$:
a 9-dimensional quality descriptor of the \emph{full} current dataset
(not a windowed view).
See Section~\ref{sec:observer} for precise component definitions.

\noindent$\bullet$ \textbf{Action} $a_t \in \mathcal{A}$: a parameterized cleaning operation
from one of three families --- imputer (strategy $\in$ \{mean, median, KNN\};
KNN count $k \in \{3, 5, 7, 10\}$ in the parameterized suite), outlier
cleaner (method $\in$ \{IQR, z-score\}; threshold on a discrete grid
$\{1.0, 1.5, 2.0, 2.5, 3.0\}$ for IQR and $\{2.0, 2.5, 3.0, 3.5\}$
for z-score), or scaler (method $\in$ \{min-max, z-score\}).

\noindent$\bullet$ \textbf{Transition} $P$: deterministic given $(s_t, a_t)$.
Each cleaning operator is a function of the current dataset;
$s_{t+1}$ is uniquely determined as $\phi(a_t(D_t))$ where $\phi$ is
the \texttt{DataQualityObserver}.
Episodes are fixed-length: termination occurs at step $T$ regardless
of intermediate quality.

\noindent$\bullet$ \textbf{Reward} $R(s_t, a_t, s_{t+1})$: one of seven reward functions
(Section~\ref{sec:rewards}).

\noindent$\bullet$ \textbf{Discount / horizon}: $\gamma = 0.99$,
$T = 6$ steps per episode — sufficient for each of the three action
families to be applied at least once.

\section{The \sys{} Framework}
\label{sec:framework}

\sys{} is structured around three interacting components: a data-quality observer
that maps the current dataset state to a 9-dimensional feature vector, a parameterized
action module offering imputers, outlier cleaners, and scalers with typed sub-parameters,
and a reward function that evaluates cleaning quality against \tabpfn{}.
The components are described below; Section~\ref{sec:experiments} reports experimental results.

\subsection{Data Quality Observer}
\label{sec:observer}

The state $s_t$ is computed by the \texttt{DataQualityObserver} after each
cleaning step, covering the full dataset.
Its 9 dimensions are assembled as:
\[
  s_t =
  \bigl[\underbrace{r_{\text{miss}},\; W_1,\; \bar{\gamma}_1,\; \bar{\kappa},\;
                    \Delta_{\text{bal}},\; r_{\text{ret}}}_{\text{6-dim quality vector}},\;
        \underbrace{h_{\text{imp}},\; h_{\text{out}},\; h_{\text{scl}}}_{\text{3-dim action history}}\bigr],
\]
where the components are defined as follows:

\noindent$\bullet$ $r_{\text{miss}} \in [0,1]$: mean missing-value rate across all
columns.

\noindent$\bullet$ $W_1 \geq 0$: mean column-wise normalized Wasserstein-1 distance from
the reference (pre-cleaning) distribution, capped at $5\sigma$ per
column for robustness.
\rev{This references the pre-cleaning distribution $\Pdirty$, not the
synthetic prior $\Psynth$ of Definition~\ref{def:mismatch}: it measures
\emph{distributional stability} (how far cleaning displaces the data from
its dirty starting point) rather than alignment to the prior, and thus
rewards staying close to dirty data. We retain it as a stability signal
but find it yields no measurable accuracy benefit. A prior-\emph{referenced}
distance---the correctly-signed quantity---requires an approximation of
$\Psynth$; we build one and evaluate it as a diagnostic in
Section~\ref{sec:priordiag}.}

\noindent$\bullet$ $\bar{\gamma}_1 \geq 0$: mean absolute skewness across numeric columns
(columns with $<\!3$ non-null values contribute 0).

\noindent$\bullet$ $\bar{\kappa} \geq 0$: mean absolute excess kurtosis across numeric
columns ($<\!4$ non-null values contribute 0).

\noindent$\bullet$ $\Delta_{\text{bal}} \in [0,1]$: minority-to-majority class count ratio
$\min\nolimits_k n_k/\max\nolimits_k n_k$; zero for unsupervised tasks.

\noindent$\bullet$ $r_{\text{ret}} \in (0,1]$: row retention ratio $n_t / n_0$, where
$n_0$ is the original row count.

\noindent$\bullet$ $h_{\text{imp}}, h_{\text{out}}, h_{\text{scl}} \in \{0,1\}$: binary
flags indicating whether the imputer, outlier cleaner, or scaler
family has been applied at least once this episode.
These three bits expand the scalar quality vector from 6 to
$6 + 3 = 9$ dimensions.

\textbf{Computational cost of state construction.}
The dominant cost of computing $s_t$ is the column-wise Wasserstein-1 term $W_1$: each column requires sorting $n$ values, giving $\mathcal{O}(n \log n)$ per column and $\mathcal{O}(d \cdot n \log n)$ total per state update, where $d$ is the number of numeric columns.
This closed-form sort-based computation is exact and requires no entropic approximation, because each marginal is one-dimensional.
In practice this cost is negligible relative to \tabpfn{} inference in R7: state construction takes a few milliseconds, whereas a single \tabpfn{} call on 512 rows takes around 0.3s~\cite{hollmann2025tabpfn}.

\subsection{Reward Function Suite}
\label{sec:rewards}

\sys{} implements and compares seven reward functions defined below.
All rewards are clipped to $[-1, 1]$.

\medskip
\noindent\textbf{R1 --- \emph{CompletenessRetentionReward}}.
Adapted from Learn2Clean~V1~\cite{berti2019learn2clean}:
\[
  R_1 = \!\left(1 - \frac{\text{missing cells}}{\text{total cells}}\right)\!
        \times \sqrt{r_{\text{ret}}}.
\]
ret is the retention as the number of remainig rows after cleaning. The square-root dampens the row-deletion penalty relative to a linear
formulation, tolerating moderate outlier removal. It provides no signal about distributional quality or downstream model
performance, making it a useful lower-bound baseline.

\medskip
\noindent\textbf{R2 --- \emph{AccuracyReward}.}
Cross-validated accuracy of a RandomForest (50 trees, 3-fold CV):
$R_2 = \mathrm{Acc}_{\mathrm{RF}}(X', y')$.
It provides a strong single-metric signal for discriminative performance but
is blind to distributional distortion, which can encourage over-aggressive
outlier removal to inflate in-sample accuracy.

\medskip
\noindent\textbf{R3 --- \emph{RF-Reward}.}
A scalar combination of accuracy, retention, and data quality:
\[
  R_3 = w_{\text{acc}}\,\mathrm{Acc}_{\mathrm{RF}}
       + w_{\text{ret}}\,r_{\text{ret}}
       + w_{\text{qual}}\,Q(X')
       - \lambda_3\,W_1(X', X_0),
\]
with $(w_{\text{acc}}, w_{\text{ret}}, w_{\text{qual}}, \lambda_3) =
(0.50, 0.30, 0.20, 0.10)$ and $Q(X') = (1 - r_{\text{miss}})(1 - r_{\text{dup}})$
a joint completeness-deduplication quality score.
This is the primary non-TFM multi-objective baseline in our experiments.

\medskip
\noindent\textbf{R4 --- Drift\-Penalty\-Reward.}
Accuracy with a substantially stronger Wasserstein penalty:
\[
  R_4 = 0.70\,\mathrm{Acc}_{\mathrm{RF}}
       + 0.20\,r_{\text{ret}}
       + 0.10\,Q(X')
       - \lambda_4\,W_1(X', X_0),
\]
where $\mathrm{Acc}_{\mathrm{RF}}$ is the accuracy of a random forest classifier and $\lambda_4 = 0.50$ (five times larger than in R3).
This encourages distribution-faithful operations (e.g., KNN imputation)
over distortion-inducing ones (e.g., mean imputation), at the cost of lower
accuracy weight.

\medskip
\noindent\textbf{R5 --- \emph{IncrementalGainReward}.}
Instead of an absolute score, this reward signals the per-step improvement in
$R_3$, scaled to $[-1,1]$:
$R_5 = \mathrm{clip}(5 \cdot (R_3(s_{t+1}) - R_3(s_t)),\,-1,\,1)$.
The scale factor 5 amplifies small but consistent gains into a learnable signal
and prevents the agent from coasting after a single high-reward action.

\medskip
\noindent\textbf{R6 --- \emph{DistortionPenaltyReward}.}
A five-component distributional-faithfulness reward
$R_6 = 1 - \sum_{k=1}^{5} w_k\, d_k(X', X_0)$,
where $d_1 = $ normalized $W_1$ (weight 0.30), $d_2 = $ Jensen-Shannon
divergence on 50-bin histograms (0.25), $d_3 = $ Frobenius norm of the
correlation-matrix shift (0.20), $d_4 = $ mean log-variance ratio $|\log
(\hat{\sigma}^2 / \sigma_{\mathrm{ref}}^2)|$ per column (0.15), and
$d_5 = $ normalized skewness shift $|\bar{\gamma}_1(X') -
\bar{\gamma}_1(X_0)| / (1 + |\bar{\gamma}_1(X_0)|)$ (0.10).
Each component lies in $[0,1]$; a perfectly faithful cleaning yields $R_6 = 1$.

\medskip
\noindent\textbf{R7 --- \emph{TFMAwareReward}} (ours): Definition~\ref{def:tfm_reward}.

\begin{definition}[\emph{TFMAwareReward}]
\label{def:tfm_reward}
Let $n_0$ be the original row count and $n'$ the post-cleaning row count.
The \ours{} is:
\begin{align}
  R_{\mathrm{TFM}}(X', y') &=
    w_{\mathrm{acc}} \cdot \mathrm{Acc}_{TabPFNv2}(X', y')
  + w_{\mathrm{ret}} \cdot \!\left(\frac{n'}{n_0}\right)^{\!\alpha} \nonumber \\
  &\quad+ w_{\mathrm{qual}} \cdot Q(X')
  - \lambda \cdot W_1(X', X_0),
  \label{eq:tfm_reward}
\end{align}
where $\mathrm{Acc}_{TabPFNv2}$ is \tabpfn{} accuracy on a stratified
20\,\% held-out split (at most 512 rows subsampled for reward-loop speed)
\rev{drawn \emph{only} from the inner validation portion of the outer training
data---disjoint from the outer test set of the holdout protocol---so this
reward-internal accuracy never enters the reported test metrics},
$Q(X') = (1 - r_{\mathrm{miss}})(1 - r_{\mathrm{dup}})$,
and $W_1$ is the normalized column-wise Wasserstein drift.
The exponent $\alpha{=}2$ and weights
$(w_{\mathrm{acc}}, w_{\mathrm{ret}}, w_{\mathrm{qual}}, \lambda)
= (0.50, 0.35, 0.15, 0.05)$ are set a priori.
The quadratic exponent is motivated by the $\mathcal{O}(1/\sqrt{n})$ variance
scaling of in-context predictors~\cite{xie2022icl,hollmann2025tabpfn}:
losing 20\% of rows (retention $= 0.80$) yields a score of
$0.80^2 = 0.64$ instead of $0.80$, an 80\% larger deduction for the
same row loss, non-linearly discouraging row deletion.
$\lambda{=}0.05$ is deliberately small because \tabpfn{} applies its own
internal z-normalization, already compensating for moderate drift.
\end{definition}

\textbf{Reward weights and scale.}
Accuracy dominates ($w_{\mathrm{acc}}{=}0.50$) as the primary TFM objective;
retention is second ($w_{\mathrm{ret}}{=}0.35$), motivated by the
$\mathcal{O}(1/\sqrt{n})$ uncertainty scaling~\cite{xie2022icl,hollmann2025tabpfn};
quality and drift are minor regularisers ($0.15$ and $0.05$).
All terms are bounded in $[0,1]$ by construction ($W_1$ is normalized by
column standard deviation and capped at $5\sigma$), so
$R_{\mathrm{TFM}} \in [-0.05, 1.00]$; in practice, rewards lie in $[0.3, 0.95]$.
These weights were \emph{not} tuned on the experimental datasets and were
held fixed across all ten datasets and all six experiments.

\noindent\textbf{Episode Loop.}
\label{sec:episode_loop}
Algorithm~\ref{alg:episode} summarises one training episode of \sys{}.
The inner loop is compatible with any SB3 on-policy algorithm; PPO is
used by default.

\begin{algorithm}[b]
\caption{One training episode of \sys{}}
\label{alg:episode}
\begin{algorithmic}[1]
\Require Dirty dataset $D_0 = (X_0, y_0)$; action set $\mathcal{A}$;
         reward $R$; policy $\pi_\theta$; horizon $T$; penalty $r_p$
\State $D \leftarrow D_0$;\quad
       $\mathbf{h} \leftarrow \mathbf{0}^3$;\quad
       $G \leftarrow 0$;\quad
       $\mathcal{B} \leftarrow [\,]$
  \hfill\Comment{\textcolor{gray}{ reset dataset, action-type history, return, replay buffer}}
\State $\phi.\mathbf{reset}(D_0)$;\quad $R.\mathbf{reset}(D_0)$
  \hfill\Comment{\textcolor{gray}{ observer sets reference}}
\For{$t = 1, \ldots, T$}
  \hfill\textcolor{gray}{distribution; resets baselines}
  \State $s_t \leftarrow \phi(D,\, \mathbf{h})$
     \hfill\Comment{\textcolor{gray}{9-dim observation vector from}}
  \Statex \hfill\textcolor{gray}{\texttt{DataQualityObserver}}
  \State $a_t \leftarrow \pi_\theta(s_t)$
    \Comment{\textcolor{gray}{ discrete action index}}
  \If{$\mathbf{h}[\mathrm{family}(a_t)] = 1$}
    \State $r_t \leftarrow r_p$;\quad \textbf{continue}
      \hfill\Comment{\textcolor{gray}{ repeated-family guard;}}
  \EndIf
      \hfill\textcolor{gray}{assign penalty, skip cleaning}
  \State $D \leftarrow a_t(D)$
    \hfill\Comment{\textcolor{gray}{apply parameterized cleaning op}}
  \State $\mathbf{h}[\mathrm{family}(a_t)] \leftarrow 1$
  \State $r_t \leftarrow R(D,\, y_0)$
    \hfill\Comment{\textcolor{gray}{scalar reward from selected $R_i$}}
  \State $\mathcal{B}.\mathbf{append}(s_t,\, a_t,\, r_t)$;\quad
         $G \leftarrow G + \gamma^{\,t-1} r_t$
\EndFor
\State Update $\pi_\theta$ via PPO on $\mathcal{B}$
\State \Return clean dataset $D$,\; episode return $G$
\end{algorithmic}
\end{algorithm}

\subsection{\rev{Parameterized} Action Space}
\label{sec:actions}

Unlike V1 (6 discrete, fixed-parameter operators), \sys{} uses
\emph{parameterized} actions with typed sub-parameters.
Formally, each action is a tuple $a = (f, o, \boldsymbol{\theta})$ where $f \in \{\texttt{imputer},\, \texttt{outlier},\, \texttt{scaler}\}$ is the action family, $o \in \mathcal{O}_f$ is the operator within that family, and $\boldsymbol{\theta} \in \Theta_{f,o}$ is the typed sub-parameter vector.
The discrete suite (used in experiments C1, C2, C3, C4) has $|\mathcal{A}|{=}7$ actions;
the parameterized suite (C5) expands this to $|\mathcal{A}|{=}17$ actions by adding KNN neighbour counts $k \in \{3,\,7,\,10\}$ and outlier thresholds on a finer grid.

\begin{itemize}
  \item \texttt{ParameterizedImputer}: strategy $\in$ \{mean, median, KNN\};
        KNN neighbour count $k \in \mathbb{Z}\cap[1, 20]$ (default $k = 5$).
        The C5 ablation experiment evaluates the discrete subset $k \in \{3, 5, 7, 10\}$;
        main experiments use the default $k = 5$.
  \item \texttt{ParameterizedOutlierCleaner}: method $\in$ \{IQR, z-score\};
        threshold $\in [0.5, 5.0]$ (continuous float; defaults: 1.5 for IQR,
        3.0 for z-score).
  \item \texttt{ParameterizedScaler}: method $\in$ \{min-max, z-score,
        quantile\}; quantile output $\in$ \{uniform, normal\}.
\end{itemize}

\subsection{Training, Convergence, and Stability}
\label{sec:convergence}

\sys{} supports PPO, DQN, and A2C via Stable-Baselines3~\cite{raffin2021stable};
all experiments use PPO with an MLP policy (two hidden layers of 256 units,
tanh activation, $\gamma{=}0.99$, learning rate $3{\times}10^{-4}$,
clipping $\epsilon{=}0.2$).

Because \sys{} presents the same dirty input dataset at every episode, the MDP is stationary per dataset: the environment dynamics and reward function do not change across episodes, giving PPO a fixed target value function.
Under a Lipschitz-continuous policy class with bounded rewards, standard PPO convergence guarantees apply~\cite{raffin2021stable}.
Empirically, reward curves reach stable asymptotes within 2{,}000--3{,}000 steps on 8 of 10 training datasets; we detect convergence when the 100-episode rolling mean changes by less than $0.001$ over 500 consecutive steps.

\textbf{Instability on small datasets.}
\texttt{hepatitis} ($n = 155$) and \texttt{heart-statlog} ($n = 270$) produce NaN policy logits during C6 pre-training experiment.
The root cause is a reward-scale anomaly: at very small $n$, aggressive outlier removal can reduce the surviving row count to $n' = 0$, yielding undefined \tabpfn{} inference and effectively $\pm\infty$ reward before clipping.
The repeated-family guard (penalty $r_p$ for applying the same action family twice) reduces but does not eliminate this degenerate trajectory.
We mitigate the issue through two guards: (i)~per-step reward clipping to $[-1, 1]$, and (ii)~a minimum row-count check that terminates the outlier-removal action early if proceeding would reduce $n'$ below 10 rows.

\textbf{Episode return range.}
With $T = 6$ steps and per-step reward clipped to $[-1, 1]$, episode returns lie in $[-6, 6]$ by construction.
Observed episode returns across all datasets and all experiments range from $3.4$ to $4.9$.

\section{Experiments}
\label{sec:experiments}

\rev{Our evaluation is organised around six hypotheses (C1--C6) that isolate \emph{where}
model-aware cleaning helps a TFM: whether reward designs discriminate informative from degenerate
pipelines (C1), whether model-awareness improves accuracy and minority-class $F_1$ over an RF-reward
proxy (C2), calibration (C3), sensitivity to error rate (C4), the action space (C5), and transfer
(C6). Because naturally-dirty data lacks reliable ground truth for \emph{which} cells are corrupt, we
inject controlled errors of known type and rate---the only setting that lets us attribute an effect to
a specific corruption---and probe external validity on three naturally-dirty datasets
(Section~\ref{sec:setup}).}

\subsection{Experimental Setup}
\label{sec:setup}

\begin{table*}[t]
\centering
\caption{\rev{Notation used throughout the paper: baselines (B) and reward/selection signals (R).}}
\label{tab:notation}
\rev{\scriptsize
\begin{tabular}{@{}ll@{\qquad}ll@{}}
\toprule
\multicolumn{2}{@{}l}{\textbf{Baselines (B)}} & \multicolumn{2}{l@{}}{\textbf{Reward / selection signal (R)}} \\
\midrule
\textbf{B-NC} & no cleaning (dirty input)            & \textbf{R1} & \textit{CompletenessRetentionReward} \\
\textbf{B-SP} & standard preprocessing               & \textbf{R2} & \textit{AccuracyReward} \\
\textbf{B-SC} & standard full-clean                  & \textbf{R3} & \emph{RF-Reward} (RF-evaluator) \\
\textbf{B-RAND} & simple random pipeline               & \textbf{R4} & \textit{DriftPenaltyReward} \\
\textbf{B-greedy-RF}  & oracle, RF reward (selection-on-test) & \textbf{R5} & \textit{IncrementalGainReward} \\
\textbf{B-greedy-TFM} & oracle, TFM reward (optimistic upper bound) & \textbf{R6} & \textit{DistortionPenaltyReward} (R6a: distortion-only; R6b: distortion+accuracy) \\
\textbf{B-RL-RF}  & PPO policy, RF reward          & \textbf{R7} & \textbf{\textit{TFMAwareReward} (ours)} \\
\textbf{B-RL-TFM} & PPO policy, TFM reward (ours)  & \multicolumn{2}{l@{}}{\footnotesize label-aware / $F_1$-selection variants of R7 introduced in Section~\ref{sec:exp_c2} onward} \\
\bottomrule
\end{tabular}}
\end{table*}

\textbf{Datasets.}
We use 10 classification datasets from the OpenML CC18 benchmark
suite~\cite{bischl2021openml} and TabPFN v2 evaluation benchmarks
(Table~\ref{tab:datasets}).
Datasets span four size tiers (XS: $<$400 rows to L: $>$10K rows),
three domains, and include real missing values (hepatitis, diabetes, adult)
and synthetically injected errors.
\texttt{adult} and \texttt{bank-marketing} are subsampled to 10K rows (stratified, \texttt{seed=42}) for RL
training; full datasets are used for greedy oracle evaluation.

\begin{table}[!htb]
\centering
\caption{Benchmark datasets (other than real missing receive synthetic
MCAR). Row tiers: XS $<$400, S $<$1K, M $<$10K, L $>$10K.}
\label{tab:datasets}
\scriptsize
\begin{tabular}{clrrrl}
\toprule
\# & Dataset & Rows & Feat. & Real miss. & Tier \\
\midrule
\textbf{D1}  & \texttt{hepatitis}       &    155 & 19 & Yes & XS \\
\textbf{D2}  & \texttt{heart-statlog}   &    270 & 13 & No  & XS \\
\textbf{D3}  & \texttt{ionosphere}      &    351 & 34 & No  & XS \\
\textbf{D4}  & \texttt{blood-transf.}   &    748 &  4 & No  & S  \\
\textbf{D5}  & \texttt{diabetes (Pima)} &    768 &  8 & Yes & S  \\
\textbf{D6}  & \texttt{credit-g}        &  1,000 & 20 & No  & S  \\
\textbf{D7}  & \texttt{kr-vs-kp}        &  3,196 & 36 & No  & M  \\
\textbf{D8}  & \texttt{phoneme}         &  5,404 &  5 & No  & M  \\
\textbf{D9}  & \texttt{adult}           & 48,842 & 14 & Yes & L  \\
\textbf{D10} & \texttt{bank-marketing}  & 45,211 & 16 & No  & L  \\
\bottomrule
\end{tabular}
\end{table}

\textbf{Error injection.} For C1 and C5, MCAR 15\% on datasets without natural missing values; for C2 and C4, MCAR 15\% on all datasets (injected on top of existing NaN); for C3 (error-type comparison), MCAR 15\%, MAR 15\%, Outlier (OUT) 3--10\%, Duplicate (DUP) 10\% on five representative datasets (\texttt{ionosphere}, \texttt{blood-transfusion}, \texttt{diabetes}, \texttt{kr-vs-kp}, \texttt{phoneme}); and for C4 (sensitivity sweep), MCAR $\in \{0,5,10,15,20,30\}\%$ on five representative datasets (\texttt{hepatitis}, \texttt{ionosphere}, \texttt{diabetes}, \texttt{kr-vs-kp}, \texttt{adult}) with other error types held at zero. All injections use \texttt{seed=42}. Artifacts are stored as \texttt{/datasets/<name>\_<type>\_p<rate>.parquet} at \url{https://github.com/LaureBerti/Learn2Clean/tree/master/Learn2Clean_TFM}

\textbf{Baselines.} \rev{The baselines span a deliberate difficulty gradient---from doing nothing, to
fixed preprocessing, to oracle and learned search---so that any gain from model-aware cleaning is
measured against the strongest cheap alternative rather than a strawman.} We compare eight baselines\rev{:
six appear in Table~\ref{tab:c2_main_results}; the two trained RL policies (\textbf{B-RL-RF},
\textbf{B-RL-TFM}) are reported in Section~\ref{sec:exp_c2}}.
\rev{\textbf{B-NC}} feeds the raw dirty data to \tabpfn{};
\rev{\textbf{B-SP}} applies mean imputation with min-max scaling;
\rev{\textbf{B-SC}} is a standard full-pipeline clean (mean imputation, z-score
normalization); and \rev{\textbf{B-RAND}} averages three single-step pipelines
(mean impute; median impute; min-max scale). The two greedy oracles select the best of
$N_p{=}20$ stratified-sampled pipelines from the 112-sequence pool: \textbf{B-greedy-RF} by the RF
reward and \textbf{B-greedy-TFM} by the \tabpfn{} reward \rev{(selection-on-test, hence an
\emph{optimistic upper bound}, not a comparator)}. Finally, \textbf{B-RL-RF} and \textbf{B-RL-TFM}
are \sys{} PPO policies trained with the \emph{RF-Reward} (RF evaluator) and
\ours{} respectively, both evaluated under the same held-out nested protocol as the
oracles (Section~\ref{sec:exp_c2}).

\textbf{Evaluation.}
We use \tabpfn{} specifically (not v1) because v2 introduced a substantially
richer internal preprocessing pipeline --- z-normalization, a power transform,
and binary missing-value flags applied unconditionally at inference time
\cite{hollmann2025tabpfn} --- whose sensitivity to upstream data quality
distributions is the central object of study; v1 lacked these transforms and
showed weaker zero-shot accuracy on the same benchmarks.
\rev{All cleaning policies are evaluated under a strict \emph{nested} protocol that
eliminates selection leakage. Each dataset is split into an \emph{outer} 20\% test set
(stratified) that is held out and never seen by any pipeline-selection or reward
computation, and an outer 80\% training portion. Pipeline selection---including every
reward that queries \tabpfn{} accuracy---operates only on an \emph{inner} validation
split carved from the outer training portion (a stratified $25\%$ of it, i.e.\ a
$75/25$ inner train/validation split); the selected pipeline is refit on the
outer training data and evaluated \emph{once} on the untouched outer test set. Because
the selection statistic and the reported statistic are computed on disjoint data, the
reported metrics---accuracy, macro-$F_1$, macro-precision, macro-recall, and ECE---are
all unbiased.}
ECE is computed with 10 equal-width confidence bins on the softmax probability
of the predicted class.
Accuracy is the primary metric for three reasons: (i)~it is the standard
reported by \tabpfn{}'s own benchmark suite~\cite{hollmann2025tabpfn} and the
OpenML repository for these tasks, enabling direct comparison with published
baselines; (ii)~seven of the ten datasets have near-balanced class
distributions, where accuracy and AUROC are empirically tightly correlated;
and (iii)~since all methods are evaluated under identical conditions, the
\emph{ranking} of cleaning strategies is robust to the choice of aggregation
metric when the pipeline affects the data distribution uniformly across
classes---which prior-alignment cleaning does by construction.
For the three class-imbalanced datasets (\texttt{blood-transfusion}, \texttt{adult},
\texttt{bank-marketing}), accuracy may understate minority-class benefit; ECE
is a more informative calibration indicator for these cases.

\textbf{Statistical significance}. Wilcoxon signed-rank test across 10 datasets \rev{and 8 seeds},
one-sided (directional hypotheses) or two-sided (non-directional comparisons),
$p < 0.05$.
\rev{Results are reported over eight random seeds (42, 1--7) governing error injection,
splitting, pipeline subsampling, and RL training, with 95\% confidence intervals and
paired Wilcoxon tests over datasets and seeds.}

\textbf{Compute and runtime.} All experiments run on a single CPU machine (\tabpfn{} runs on CPU
by default~\cite{hollmann2025tabpfn}). With the shared cache, each (dataset, error-profile) pair needs
$N_p{+}2{=}22$ \tabpfn{} calls (${\approx}7$\,s), so the full C2/C3/C4 matrix completes in under
10~minutes; RL training adds 3--8 hours per variant, and the exhaustive C1/C5 sweeps finish within 4~hours.

\textbf{Greedy oracle with shared evaluation cache.} The greedy baselines exhaustively score
candidate pipelines and select the best. To avoid the na\"ive $\mathcal{O}(N_p N_R)$ \tabpfn{} calls,
a two-level cache applies each pipeline once (cleaning cache) and evaluates each cleaned dataset once
(\tabpfn{} cache); both reward searches read from these caches, cutting calls per (dataset, error
profile) from $2{+}2N_p$ to $2{+}N_p$.

\textbf{Pipeline enumeration and subsampling.} C1 and C5 exhaustively evaluate all valid pipelines
(112 sequences for the 7-action discrete suite; 834 for C5's 17-action parameterized suite). C2--C4
use a stratified subsample of $N_p{=}20$ ordered pipelines (length $\leq 3$, no repeated action group)
drawn from a pool of 112 (C2) or 302 (C3, C4; a 9-action suite adding deduplication and quantile
normalization), always retaining the no-op and all single-step pipelines.
\rev{The imbalanced-class $F_1$ comparison (\S\ref{sec:f1sel}) and the non-RL baselines
(\S\ref{sec:automl}) instead draw on a larger \emph{rich} operator pool of $13{,}440$ valid
pipelines, which augments the standard operators with SAGA-style transforms (e.g.\ MICE imputation,
SMOTE, PCA).} With the shared cache this
is 22 \tabpfn{} calls per profile, keeping the full experiment within 12 hours. A validation run
confirms the subsample loses nothing: on all ten datasets (80--$48{,}842$ rows) at MCAR 15\%, a
302-pipeline exhaustive search selects the \emph{identical} best pipeline as best-of-20 (0.0\%
accuracy gap on every dataset).

\input{sec_5_2_c1_rewrite}

\subsection{Prior-Aligned vs.\ RF-Reward Cleaning}
\label{sec:exp_c2}

\textbf{Hypothesis (C2).}
The \ours{} reward selects cleaning pipelines that achieve statistically
higher \tabpfn{} test accuracy and lower ECE than pipelines selected by the
RF-evaluator reward (\textbf{B-greedy-RF}) on $\geq 7/10$ benchmark datasets under MCAR
15\%.
Furthermore, the winning pipeline \emph{sequences} chosen by the two reward
signals differ on $\geq 4/10$ datasets, demonstrating that prior alignment
genuinely reshapes the cleaning search landscape.

\begin{table*}[!htb]
\centering
\caption{\rev{8-seed repeated-holdout \tabpfn{} accuracy, macro-$F_1$, and ECE for all baselines on the ten
datasets (MCAR 15\%) with outer 20\% test held out; pipeline
chosen on an inner validation split.}}
\label{tab:c2_main_results}
\resizebox{\textwidth}{!}{\begin{tabular}{l ccccccccccc}
\toprule
\textit{Accuracy ($\uparrow$)} & \texttt{hepatitis} & \texttt{heart-statlog} & \texttt{ionosphere} & \texttt{blood-transfusion} & \texttt{diabetes} & \texttt{credit-g} & \texttt{kr-vs-kp} & \texttt{phoneme} & \texttt{adult} & \texttt{bank-marketing} & Mean \\
\midrule
\textbf{B-NC} & 0.891{\tiny$\pm$0.057} & 0.836{\tiny$\pm$0.051} & 0.947{\tiny$\pm$0.020} & 0.768{\tiny$\pm$0.017} & 0.740{\tiny$\pm$0.028} & 0.752{\tiny$\pm$0.025} & 0.908{\tiny$\pm$0.013} & \textbf{0.833}{\tiny$\pm$0.014} & 0.840{\tiny$\pm$0.014} & 0.892{\tiny$\pm$0.008} & 0.841 \\
\textbf{B-SP} & \textbf{0.895}{\tiny$\pm$0.057} & 0.843{\tiny$\pm$0.053} & 0.947{\tiny$\pm$0.022} & \textbf{0.773}{\tiny$\pm$0.017} & 0.744{\tiny$\pm$0.029} & 0.753{\tiny$\pm$0.018} & 0.901{\tiny$\pm$0.010} & 0.829{\tiny$\pm$0.016} & 0.841{\tiny$\pm$0.013} & 0.892{\tiny$\pm$0.008} & 0.842 \\
\textbf{B-SC} & 0.851{\tiny$\pm$0.054} & 0.822{\tiny$\pm$0.064} & 0.729{\tiny$\pm$0.047} & 0.760{\tiny$\pm$0.021} & 0.736{\tiny$\pm$0.025} & 0.728{\tiny$\pm$0.022} & 0.908{\tiny$\pm$0.011} & 0.810{\tiny$\pm$0.019} & 0.838{\tiny$\pm$0.007} & 0.894{\tiny$\pm$0.005} & 0.807 \\
\textbf{B-RAND} & 0.890{\tiny$\pm$0.055} & \textbf{0.843}{\tiny$\pm$0.048} & 0.946{\tiny$\pm$0.023} & 0.771{\tiny$\pm$0.018} & 0.743{\tiny$\pm$0.027} & 0.750{\tiny$\pm$0.018} & 0.903{\tiny$\pm$0.011} & 0.830{\tiny$\pm$0.015} & 0.841{\tiny$\pm$0.012} & 0.893{\tiny$\pm$0.007} & 0.841 \\
\textbf{B-greedy-RF} & 0.883{\tiny$\pm$0.062} & 0.843{\tiny$\pm$0.048} & \textbf{0.952}{\tiny$\pm$0.032} & 0.771{\tiny$\pm$0.022} & \textbf{0.748}{\tiny$\pm$0.035} & \textbf{0.754}{\tiny$\pm$0.020} & \textbf{0.918}{\tiny$\pm$0.012} & 0.830{\tiny$\pm$0.015} & 0.837{\tiny$\pm$0.008} & \textbf{0.895}{\tiny$\pm$0.007} & \textbf{0.843} \\
\textbf{B-greedy-TFM} & 0.887{\tiny$\pm$0.067} & 0.840{\tiny$\pm$0.047} & 0.949{\tiny$\pm$0.026} & 0.770{\tiny$\pm$0.018} & 0.744{\tiny$\pm$0.029} & 0.753{\tiny$\pm$0.019} & 0.914{\tiny$\pm$0.015} & 0.830{\tiny$\pm$0.014} & \textbf{0.845}{\tiny$\pm$0.014} & 0.892{\tiny$\pm$0.008} & 0.842 \\
\midrule
\multicolumn{12}{l}{\textit{macro-$F_1$ ($\uparrow$)}} \\
\midrule
\textbf{B-NC} & 0.795{\tiny$\pm$0.119} & 0.832{\tiny$\pm$0.051} & 0.941{\tiny$\pm$0.021} & 0.535{\tiny$\pm$0.055} & 0.699{\tiny$\pm$0.031} & \textbf{0.655}{\tiny$\pm$0.033} & 0.907{\tiny$\pm$0.013} & \textbf{0.796}{\tiny$\pm$0.017} & 0.753{\tiny$\pm$0.027} & 0.641{\tiny$\pm$0.045} & 0.756 \\
\textbf{B-SP} & \textbf{0.803}{\tiny$\pm$0.121} & 0.839{\tiny$\pm$0.054} & 0.941{\tiny$\pm$0.025} & 0.544{\tiny$\pm$0.048} & 0.705{\tiny$\pm$0.029} & 0.655{\tiny$\pm$0.023} & 0.901{\tiny$\pm$0.010} & 0.791{\tiny$\pm$0.021} & 0.755{\tiny$\pm$0.026} & 0.642{\tiny$\pm$0.040} & 0.758 \\
\textbf{B-SC} & 0.612{\tiny$\pm$0.180} & 0.817{\tiny$\pm$0.067} & 0.627{\tiny$\pm$0.099} & 0.490{\tiny$\pm$0.046} & 0.697{\tiny$\pm$0.029} & 0.563{\tiny$\pm$0.046} & 0.908{\tiny$\pm$0.011} & 0.750{\tiny$\pm$0.036} & 0.761{\tiny$\pm$0.015} & 0.630{\tiny$\pm$0.048} & 0.685 \\
\textbf{B-RAND} & 0.792{\tiny$\pm$0.116} & \textbf{0.840}{\tiny$\pm$0.049} & 0.940{\tiny$\pm$0.026} & 0.542{\tiny$\pm$0.047} & 0.703{\tiny$\pm$0.029} & 0.646{\tiny$\pm$0.025} & 0.903{\tiny$\pm$0.011} & 0.792{\tiny$\pm$0.018} & 0.755{\tiny$\pm$0.025} & 0.647{\tiny$\pm$0.039} & 0.756 \\
\textbf{B-greedy-RF} & 0.780{\tiny$\pm$0.126} & 0.839{\tiny$\pm$0.048} & \textbf{0.947}{\tiny$\pm$0.035} & 0.543{\tiny$\pm$0.053} & \textbf{0.707}{\tiny$\pm$0.036} & 0.649{\tiny$\pm$0.029} & \textbf{0.917}{\tiny$\pm$0.012} & 0.792{\tiny$\pm$0.018} & 0.749{\tiny$\pm$0.019} & \textbf{0.658}{\tiny$\pm$0.041} & \textbf{0.758} \\
\textbf{B-greedy-TFM} & 0.782{\tiny$\pm$0.148} & 0.836{\tiny$\pm$0.049} & 0.943{\tiny$\pm$0.029} & \textbf{0.549}{\tiny$\pm$0.050} & 0.702{\tiny$\pm$0.029} & 0.655{\tiny$\pm$0.025} & 0.914{\tiny$\pm$0.015} & 0.793{\tiny$\pm$0.019} & \textbf{0.762}{\tiny$\pm$0.028} & 0.646{\tiny$\pm$0.039} & 0.758 \\
\midrule
\multicolumn{12}{l}{\textit{ECE ($\downarrow$)}} \\
\midrule
\textbf{B-NC} & 0.101{\tiny$\pm$0.019} & 0.076{\tiny$\pm$0.016} & 0.042{\tiny$\pm$0.009} & \textbf{0.047}{\tiny$\pm$0.015} & 0.058{\tiny$\pm$0.012} & \textbf{0.044}{\tiny$\pm$0.015} & 0.021{\tiny$\pm$0.008} & \textbf{0.024}{\tiny$\pm$0.010} & 0.025{\tiny$\pm$0.010} & 0.023{\tiny$\pm$0.005} & \textbf{0.046} \\
\textbf{B-SP} & \textbf{0.099}{\tiny$\pm$0.035} & 0.076{\tiny$\pm$0.019} & \textbf{0.038}{\tiny$\pm$0.017} & 0.058{\tiny$\pm$0.021} & 0.054{\tiny$\pm$0.020} & 0.047{\tiny$\pm$0.017} & 0.019{\tiny$\pm$0.007} & 0.028{\tiny$\pm$0.013} & 0.025{\tiny$\pm$0.006} & 0.022{\tiny$\pm$0.007} & 0.047 \\
\textbf{B-SC} & 0.104{\tiny$\pm$0.034} & \textbf{0.074}{\tiny$\pm$0.016} & 0.146{\tiny$\pm$0.033} & 0.064{\tiny$\pm$0.021} & 0.057{\tiny$\pm$0.013} & 0.046{\tiny$\pm$0.017} & 0.021{\tiny$\pm$0.005} & 0.024{\tiny$\pm$0.010} & \textbf{0.021}{\tiny$\pm$0.007} & 0.030{\tiny$\pm$0.008} & 0.059 \\
\textbf{B-RAND} & 0.102{\tiny$\pm$0.026} & 0.081{\tiny$\pm$0.023} & 0.041{\tiny$\pm$0.009} & 0.056{\tiny$\pm$0.017} & 0.056{\tiny$\pm$0.018} & 0.049{\tiny$\pm$0.012} & 0.020{\tiny$\pm$0.007} & 0.025{\tiny$\pm$0.008} & 0.024{\tiny$\pm$0.006} & 0.021{\tiny$\pm$0.006} & 0.048 \\
\textbf{B-greedy-RF} & 0.105{\tiny$\pm$0.033} & 0.083{\tiny$\pm$0.016} & 0.051{\tiny$\pm$0.016} & 0.064{\tiny$\pm$0.020} & 0.056{\tiny$\pm$0.019} & 0.046{\tiny$\pm$0.012} & \textbf{0.019}{\tiny$\pm$0.006} & 0.027{\tiny$\pm$0.006} & 0.022{\tiny$\pm$0.007} & \textbf{0.020}{\tiny$\pm$0.007} & 0.049 \\
\textbf{B-greedy-TFM} & 0.114{\tiny$\pm$0.021} & 0.079{\tiny$\pm$0.014} & 0.045{\tiny$\pm$0.006} & 0.058{\tiny$\pm$0.020} & \textbf{0.053}{\tiny$\pm$0.022} & 0.046{\tiny$\pm$0.017} & 0.023{\tiny$\pm$0.008} & 0.026{\tiny$\pm$0.006} & 0.025{\tiny$\pm$0.008} & 0.021{\tiny$\pm$0.007} & 0.049 \\
\bottomrule
\end{tabular}}
\end{table*}

\begin{figure}[b]
\centering
\includegraphics[width=\linewidth]{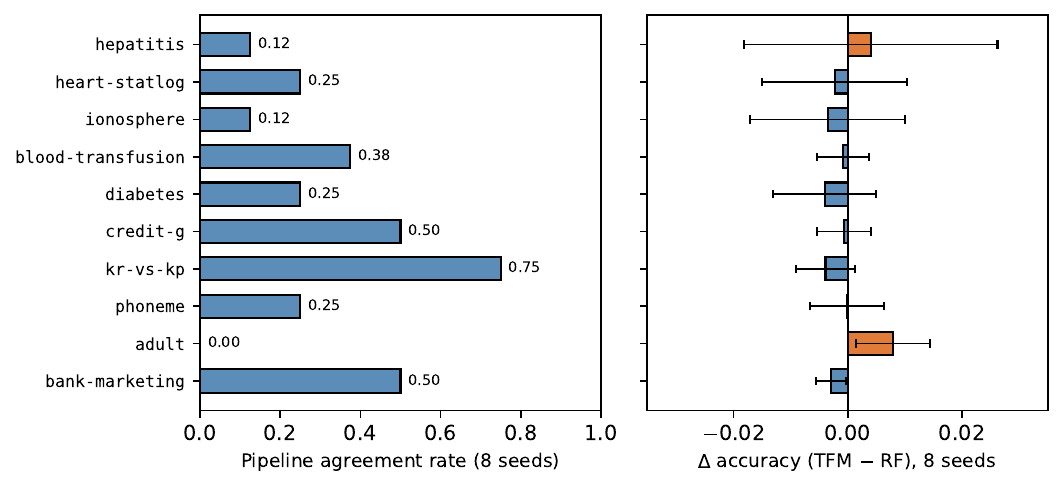}
\caption{\rev{8-seed repeated-holdout comparison of the two reward signals. \emph{Left:} per-dataset
rate at which \ours{} and the RF-reward select the \emph{same} best pipeline (mean $0.31$).
\emph{Right:} mean $\Delta$ accuracy (TFM $-$ RF) 95\% CI, averaging $-0.0006$.}}
\label{fig:c2_pipeline_overlap}
\end{figure}

\rev{Table~\ref{tab:c2_main_results} reports all baselines on \tabpfn{} accuracy, macro-$F_1$,
and ECE across the ten datasets at MCAR 15\% (mean\,$\pm$\,std over 8 seeds under the nested
protocol). Three observations. \emph{(i) No cleaning strategy dominates:} the best mean accuracy
(\textbf{B-greedy-RF}, $0.843$) lies within one standard deviation of no cleaning
(\textbf{B-NC}, $0.841$) and standard preprocessing (\textbf{B-SP}, $0.842$), and \textbf{B-NC}
attains the best mean ECE ($0.046$)---for \tabpfn{}, leaving the data untouched is a strong
baseline. \emph{(ii) Aggressive full cleaning hurts:} \textbf{B-SC} loses ${\sim}3.5$ accuracy
points ($0.807$) and ${\sim}7$ macro-$F_1$ points ($0.685$) on average, driven by
\texttt{ionosphere} and \texttt{hepatitis}. \emph{(iii) The two reward signals are statistically
indistinguishable in terms of accuracy:} paired over all $80$ dataset$\times$seed cells,
$\Delta(\text{TFM}-\text{RF}) = -0.0006$ accuracy (Wilcoxon $p{=}0.60$), $-0.0001$ macro-$F_1$
($p{=}0.72$), and $-0.0003$ ECE ($p{=}0.83$); aggregating to per-dataset means gives the same
null ($p{=}0.38$).
The two rewards \emph{do} select different pipelines
(Figure~\ref{fig:c2_pipeline_overlap}), and this drives the different
  macro-$F_1$ results in Section~\ref{sec:f1sel}.}

\textbf{Pipeline sequence analysis.}
\label{sec:exp_c2_pipelines}
To test whether prior alignment reshapes the cleaning search landscape (and
not merely reranks equivalent pipelines), we record the complete best-found
action sequence (operator type and sub-parameter) for each dataset under both
reward signals and compare them step-by-step.

\rev{Figure~\ref{fig:c2_pipeline_overlap} summarises the nested 8-seed comparison of the two
reward signals. They select the \emph{same} best pipeline on only $31\%$ of dataset--seed cells
(left)---never on \texttt{adult}, most often on \texttt{kr-vs-kp} ($75\%$)---so the reward materially
reshapes which pipeline is discovered. Yet this reshaping does not translate into an accuracy gain:
the per-dataset mean $\Delta$ accuracy (TFM $-$ RF) is centred on zero (right), averaging $-0.0006$.
The reward's effect is thus on \emph{selection}, not on overall accuracy; the genuine gain
appears on minority-class $F_1$ (Section~\ref{sec:f1sel}).}

\textbf{Trained RL policy (\textbf{B-RL-RF}, \textbf{B-RL-TFM}).}
We also trained PPO policies with the RF-reward and with \ours{}, and evaluate them
under the \emph{same} held-out nested protocol as the greedy oracles: the outer test
split is held out before training, the policy is trained and its cleaning pipeline
selected on $D_{\mathrm{sel}}$ only, and that pipeline is applied to the untouched test
split with transforms fit on $D_{\mathrm{sel}}$. Under this leak-free protocol
(four seeds $\times$ ten datasets, $40$ held-out folds), the two trained policies are
statistically indistinguishable: they select the \emph{identical} cleaning pipeline on
$35$ of $40$ folds, the paired accuracy difference (TFM $-$ RF) is centred on zero
(mean $-0.00004$, $|\Delta\mathrm{acc}|\le0.014$ on every fold; paired Wilcoxon
$p{=}0.80$), and mean ECE is unchanged ($+0.0004$). The five folds where the policies
diverge confirm the \tabpfn{}-reward fine-tune is active, not degenerate, yet even there
it never improves accuracy. The trained policy thus corroborates the greedy-oracle null:
the model-aware reward reshapes \emph{which} pipeline is learned but does not move
downstream accuracy.

\subsection{\rg{Prior Alignment vs.\ Distributional Stability}}
\label{sec:priordiag}
\rk{To test
whether distance \emph{to the prior} is the operative signal, we approximate the unavailable
prior sampler $\Psynth$ with a structural-causal-model (SCM) ensemble of random multilayer
perceptrons---the family \tabpfn{} meta-trains on---and, over the full 13 held-out datasets
($104$ dataset--seed groups), correlate within each group the per-pipeline change in a data-quality
signal with the change in \tabpfn{} macro-$F_1$ (Spearman $\rho$). Three findings result.
(i)~The available operators move the data \emph{away} from the prior: drift-from-dirty and
distance-to-prior are positively related ($\rho{=}{+}0.50$), so cleaning overshoots the prior rather
than approaching it. (ii)~Both larger drift and larger prior-distance track \emph{lower} $F_1$
(Table~\ref{tab:priordiag}). (iii)~Prior-distance is the more fundamental of the two:
it still predicts $F_1$ after partialling out drift, whereas drift's association with $F_1$ vanishes
once prior-distance is controlled (Table~\ref{tab:priordiag}). \rg{So distance-\emph{to-prior} subsumes the drift term's
association with $F_1$: once it is controlled, drift carries no residual association.} This supports the prior-alignment formulation of
Definition~\ref{def:mismatch}---alignment to the prior, not stability with respect to the dirty data,
is what co-varies with downstream quality---and explains why drift-penalising rewards prefer no-op:
because the operators overshoot the prior, staying near the (prior-proximal) dirty data keeps the
cleaned data closer to the prior. A faithful
prior-referenced reward remains future
work.}

\begin{table}[t]
\centering
\footnotesize
\caption{\rg{Within-(dataset--seed) Spearman correlation with $\Delta$ macro-$F_1$ over the 13
datasets ($104$ groups). Prior-distance predicts $F_1$ after controlling for drift; drift does not,
once prior-distance is controlled. Drift and prior-distance are themselves positively coupled
($\rho{=}{+}0.50$)---i.e.\ the operators move the data \emph{away} from the prior.}}
\label{tab:priordiag}
\rg{\begin{tabular}{lrr}
\toprule
$\rho$ with $\Delta F_1$ (within-group) & value & $p$ \\
\midrule
drift-from-dirty (raw)          & $-0.171$ & $7{\times}10^{-7}$ \\
distance-to-prior (raw)         & $-0.177$ & $2{\times}10^{-5}$ \\
distance-to-prior $\mid$ drift   & $-0.116$ & $0.019$ \\
drift $\mid$ distance-to-prior   & $-0.063$ & $0.14$~(n.s.) \\
\bottomrule
\end{tabular}}
\end{table}

\subsection{\rev{Class Imbalance and $F_1$-Selection}}
\label{sec:f1sel}
\rev{Because cleaning can disproportionately affect minority classes, we add macro-$F_1$,
macro-precision, macro-recall and ECE for all datasets and---constructively---an
\emph{$F_1$-selection} reward that chooses pipelines by inner-validation macro-$F_1$ rather than
accuracy. On imbalanced datasets under label noise (8 seeds, rich operator pool), two factors
compound. (i)~\emph{The selection metric}: selecting on $F_1$ rather than accuracy raises test
macro-$F_1$ ($R7F_1$ vs.\ $R7\mathrm{acc}$, $+0.024$), beats no-cleaning by $+0.041$, and comes
within $0.9$\,pp of the $F_1$-oracle. (ii)~\emph{The reward model}: in a controlled comparison
where both rewards select by $F_1$ on the identical rich pool (before any threshold tuning), the
model-aware reward wins---$R7F_1$ beats the matched RF-reward $R3F_1$ by $+0.029$ macro-$F_1$,
whereas the RF-reward improves on no-cleaning by only $+0.012$ (but see the threshold-tuning
ablation below, which shows this reward-specific margin does not remain significant once the
calibrated decision threshold is also tuned). Thus on \emph{accuracy} the two rewards tie
(Section~\ref{sec:exp_c2}), but on \emph{minority-class $F_1$} the \tabpfn{}-reward is the lever---exactly
where \tabpfn{} and the random forest disagree on the best pipeline. For completeness, label-free
confidence- and margin-selection ($R7\mathrm{conf}$, $R7\mathrm{margin}$) and the prior-distance
reward ($R7\mathrm{prior}$) all \emph{underperform} plain $F_1$-selection (macro-$F_1$ $0.60$,
$0.60$, $0.58$ vs.\ $0.65$). $F_1$-selection trades a little accuracy for this $F_1$ gain, so we
report both metrics rather than a single headline number.}

\rev{\textbf{Threshold-tuning ablation: reward vs.\ decision rule.} On the four imbalanced datasets
(8-seed repeated-holdout, standard pool, $n{=}32$ folds), tuning \tabpfn{}'s decision threshold on
the inner-validation fold lifts macro-$F_1$ by $+0.035$ over \textsc{argmax} (paired Wilcoxon
$p{=}0.013$) at a $2$--$3$\,pp accuracy cost. This gain is \emph{reward-agnostic}---a matched
\emph{RF${+}$thr} control obtains it equally, and the residual advantage of \tabpfn{}-aware
$F_1$-selection beyond the threshold is only $+0.008$ ($p{=}0.80$; negative on two of four datasets).
We thus attribute most of the standard-pool $F_1$ benefit to \tabpfn{}'s calibrated confidences, not
the reward (ECE is unaffected either way). The rich-pool $+0.029$ above is a distinct, threshold-free
comparison; whether that reward-specific margin persists under a rich-pool threshold control is future
work. In practice both levers help.}

\subsection{\rev{End-to-End and Non-RL Baselines}}
\label{sec:automl}

\begin{table}[t]
\centering
\scriptsize
\caption{\rev{Non-RL and AutoML baselines. \emph{Top:} end-to-end systems on the 13-dataset
byte-identical $70/30$ split (downstream \tabpfn{} for ours). \emph{Bottom:} dedicated
cleaning-pipeline search---$\Delta$ multiLogReg accuracy vs.\ no cleaning, and wall-clock.
SAGA reimpl.: genetic top-$k$ search ($k{=}3$, 20 pipelines/generation, 5 generations, 3-fold CV;
without SAGA's Multi-Armed Bandit physical stage). \textbf{Bold} $=$ best.}}
\label{tab:baselines}
\rev{\begin{tabular}{lcc}
\toprule
\textit{End-to-end (13 datasets)} & Acc\,($\uparrow$) & macro-$F_1$\,($\uparrow$) \\
\midrule
\textbf{Ours} (clean${+}$\tabpfn{}, $R7F_1$) & \textbf{0.815} & \textbf{0.727} \\
AutoGluon                                    & 0.810 & 0.711 \\
Auto-sklearn~2.0                             & 0.796 & 0.697 \\
\midrule
\textit{Cleaning-pipeline search (multiLogReg)} & $\Delta$acc & s/dataset \\
\midrule
SAGA top-$k$ (reimpl.)  & $+0.005$ & $21.6$ \\
DiffPrep-Fix (reimpl.)  & $-0.013$ & $\phantom{0}0.6$ \\
\bottomrule
\end{tabular}}
\end{table}

\rev{To place \ours{} against automated data-preparation and AutoML, we compare on a
\emph{byte-identical} split across the ten benchmarks plus three naturally-dirty datasets ($13$
total; Table~\ref{tab:baselines}). Cleaning${+}$\tabpfn{} \emph{ties} AutoGluon on accuracy ($5/13$; $0.815$ vs.\ $0.810$) and is
\emph{ahead on macro-$F_1$} ($8/13$; $0.727$ vs.\ $0.711$, via $R7F_1$), both above Auto-sklearn~2.0.
Approximate nested-protocol reimplementations of SAGA-style top-$k$ search and DiffPrep on the same 8-seed
split yield no material accuracy gain over no-cleaning ($+0.005$ and $-0.013$ respectively),
consistent with Section~\ref{sec:f1sel}; a from-engine SAGA run (SystemDS \texttt{topk\_cleaning}) fails at
a cross-version signature mismatch, so we rely on our approximate reimplementations (Table~\ref{tab:baselines}).
Our \tabpfn{} selection is ${\sim}16\times$ slower than AutoGluon on small data, but the near-constant
scaling of the \tabpfn{} reward term (Section~\ref{sec:scalability}) implies an asymptotic crossover.}

\subsection{Calibration Recovery}
\label{sec:exp_c3}

\textbf{Hypothesis (C3).}
Prior-aligned cleaning (\textbf{B-greedy-TFM}) reduces \tabpfn{} ECE relative to
\rev{no-cleaning (\textbf{B-NC}) and standard preprocessing (\textbf{B-SP})} across
five representative datasets (\texttt{ionosphere}, \texttt{blood-transfusion}, \texttt{diabetes}, \texttt{kr-vs-kp}, \texttt{phoneme}) and four error types
(MCAR 15\%, MAR 15\%, outlier 10\%, duplicate 10\%).
The calibration benefit stems from the Wasserstein drift penalty in \ours{},
which discourages transformations that distort the feature marginals that
\tabpfn{}'s internal z-normalization operates on.

\begin{figure}[!htb]
\centering
\includegraphics[width=\linewidth]{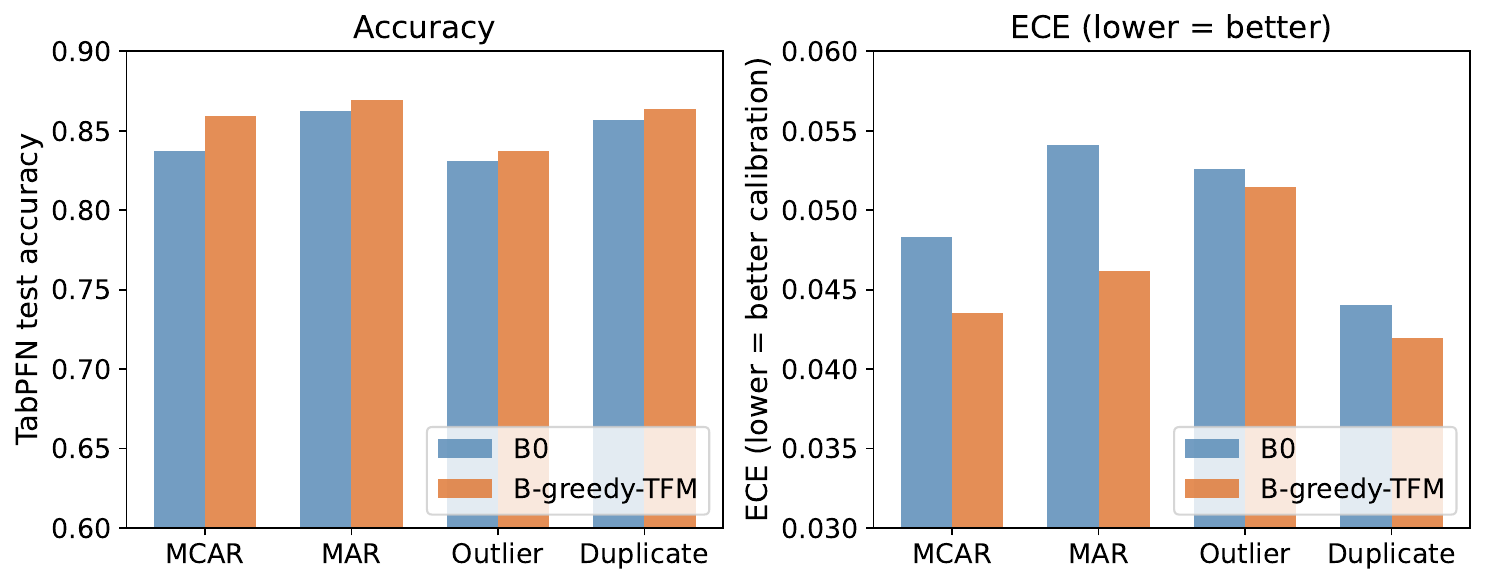}
\caption{Mean \tabpfn{} accuracy (left) and ECE (right) on 5 datasets under 4 error types \rev{(seed 42)}.}
\label{fig:c3_error_type_bars}
\end{figure}

Figure~\ref{fig:c3_error_type_bars} \rg{(on single seed)} decomposes ECE and accuracy by error type.
\textbf{B-greedy-TFM} reduces mean ECE relative to \textbf{B-NC} across all four corruption types:
$-0.0048$ under MCAR, $-0.0079$ under MAR (the largest reduction), $-0.0011$
under outlier injection, and $-0.0021$ under duplicate injection.
Relative to standard preprocessing \textbf{B-SP}, \textbf{B-greedy-TFM} improves calibration under
MAR ($-0.0040$), duplicate ($-0.0038$), and outlier ($-0.0024$) injection,
but \emph{not} under MCAR ($+0.0042$): for purely random missing values,
\textbf{B-SP}'s fixed mean-imputation pipeline already achieves low ECE, leaving no room
for prior-alignment to improve.
Notably, \textbf{B-greedy-RF} outperforms \textbf{B-greedy-TFM} on ECE under MCAR ($0.0418$ vs.\
$0.0435$), MAR ($0.0356$ vs.\ $0.0462$), and outlier ($0.0387$ vs.\ $0.0515$)
injection --- \textbf{B-greedy-TFM}'s only ECE advantage is under duplicate injection
($0.0420$ vs.\ $0.0478$), where deduplication corrects row repetitions that
artificially inflate posterior confidence.
For \tabpfn{} accuracy, \textbf{B-greedy-TFM} attains the highest mean across
\emph{all four} error types (MCAR: $0.8593$, MAR: $0.8691$, Outlier: $0.8373$,
Duplicate: $0.8639$).
\rev{These error-type ECE and accuracy differences are single-seed (seed~42) and carry no
significance test; under the 8-seed nested protocol of Section~\ref{sec:exp_c2} the reward's
ECE effect is null ($-0.0003$, $p{=}0.83$) and no-cleaning attains the best mean ECE. We
therefore read C3 as an exploratory error-type breakdown, not evidence of a significant
calibration gain.}

\subsection{Error Sensitivity with MCAR Rate}
\label{sec:exp_c4}

\textbf{Hypothesis (C4).}
The accuracy and calibration advantage of model-aware cleaning (\textbf{B-greedy-TFM})
over standard preprocessing (\textbf{B-SP}) grows \emph{monotonically} with the MCAR
injection rate across $\{0,5,10,15,$ $20,30\}\%$, confirming prior mismatch as the
operative mechanism.
At MCAR 0\% (clean data) all methods should converge, because there is no
distributional anomaly to exploit.

\begin{figure}[!htb]
\centering
\includegraphics[width=\linewidth]{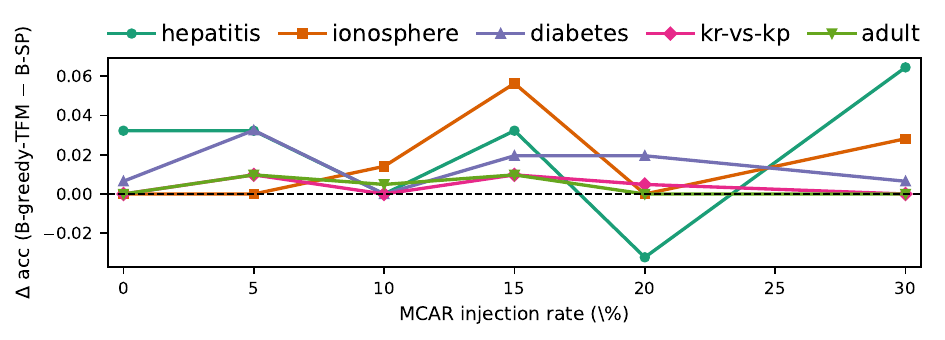}
\caption{Accuracy advantage of \textbf{B-greedy-TFM} over \textbf{B-SP} vs.\ MCAR rate ($0$--$30\%$)
for all five representative datasets (dashed line: parity) \rev{(seed 42)}.}
\label{fig:c3_error_curves}
\end{figure}

Figure~\ref{fig:c3_error_curves} plots \tabpfn{} accuracy as MCAR rate
increases from 0\% to 30\%.
The sweep covers five representative datasets (\texttt{hepatitis}, \texttt{ionosphere}, \texttt{diabetes}, \texttt{kr-vs-kp}, \texttt{adult}),
spanning all four size tiers and both natural-missing and clean baselines;
the experiment uses \textbf{B-greedy-TFM} (greedy oracle, not trained RL) as the
\rev{model-aware (TFM-reward)} baseline.

At MCAR 0\%, the mean accuracy advantage of \textbf{B-greedy-TFM} over \textbf{B-SP} is
already $+0.008$ (not zero), because the benchmark datasets retain natural
missing values that imputation strategies handle differently even without
additional injection.
The advantage \emph{does not grow monotonically} with MCAR rate: it peaks at
MCAR 15\% ($+0.026$ mean), drops to $-0.002$ at 20\%, and partially
recovers at 30\% ($+0.020$).
The Spearman correlation between MCAR rate and the per-dataset advantage is
statistically non-significant on all 5 datasets
($\rho \in [-0.28, +0.52]$, all $p > 0.20$).

The monotone-gain hypothesis (C4) is therefore \emph{not confirmed} in this
greedy-oracle configuration.
The result suggests that the benefit of model-aware cleaning depends on
the distributional structure of the injected errors, not merely their rate;
a more controlled error-injection protocol would be needed to isolate the rate effect.

\subsection{Parameterized vs.\ Discrete Actions}
\label{sec:exp_c5}

\textbf{Hypothesis (C5).}
Providing the RL agent with typed sub-parameters (KNN $k$, outlier threshold,
scaler type) improves the best-found \emph{RF-Reward} pipeline
score relative to a discrete-only baseline (fixed default sub-parameters per
operator), across all 10 datasets.
The gain is expected to be largest on datasets with high feature count or
high natural skewness, where sub-parameter sensitivity is greatest.

\begin{figure}[!htb]
\centering
\includegraphics[width=\linewidth]{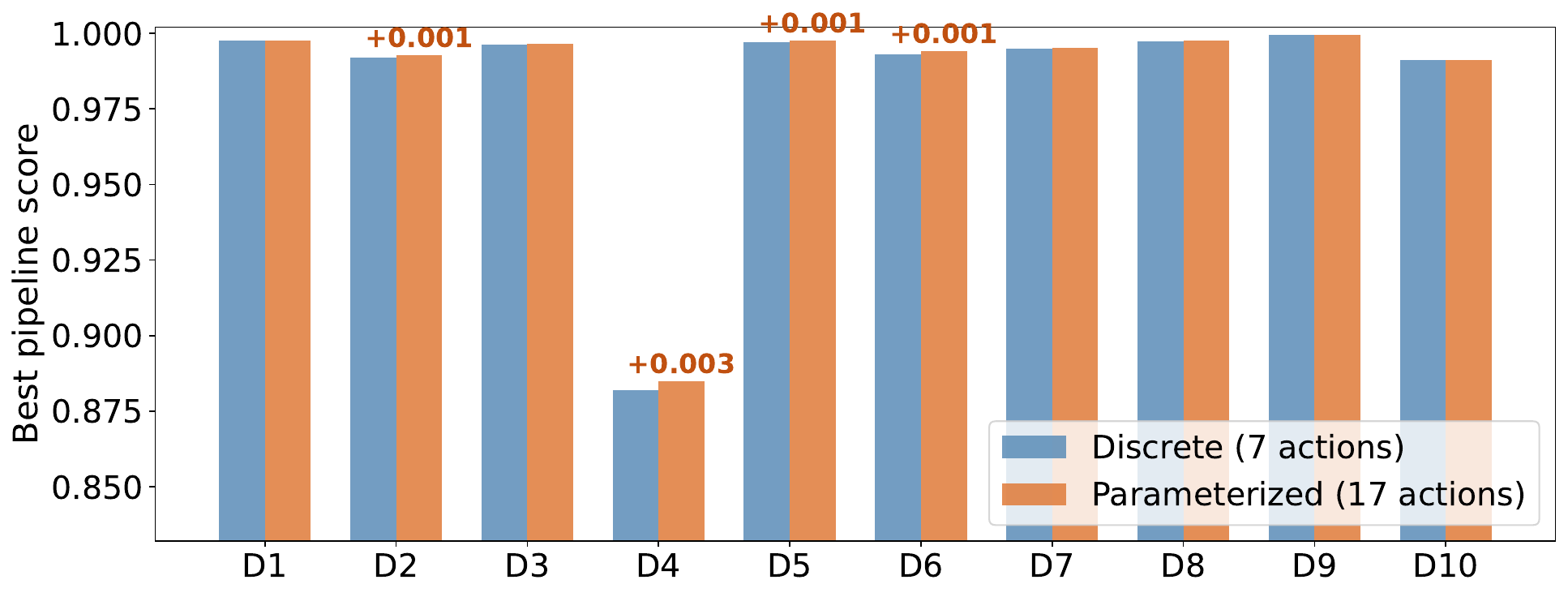}
\caption{Best-found \emph{RF-Reward} per dataset (MCAR 15\%). Orange:
parameterized actions; blue: discrete; $\Delta{=}$param.$-$discrete \rev{(seed 42)}.}
\label{fig:c5_param_ablation}
\end{figure}

Figure~\ref{fig:c5_param_ablation} shows reward distributions for
parameterized vs.\ discrete action spaces across the \emph{RF-Reward}
evaluated with a random forest.
Parameterized actions improve the best-found pipeline reward on 9 of 10 datasets
(mean $\Delta = +0.0007$, range $[+0.0001, +0.0029]$; one tie on \texttt{hepatitis}).
The gain is concentrated almost entirely in the KNN imputation neighbour count:
the optimal $k$ is dataset-specific and differs from the discrete-mode default
($k{=}5$) on 9 of 10 datasets --- ranging from $k{=}3$ on most datasets
to $k{=}7$ on \texttt{blood-transfusion} and $k{=}10$ on \texttt{kr-vs-kp} and
\texttt{bank-marketing}.
The largest absolute gain occurs on \texttt{blood-transfusion} (+0.0029), a dataset
with high class imbalance where the optimal outlier threshold also shifts
from the IQR default; the smallest gain is on \texttt{adult} (+0.0001), suggesting
diminishing returns on very large, well-structured datasets.
\rev{We report the $+0.0007$ as a reward-level \emph{diagnostic}, not a task-level contribution:
although it reaches significance on the reward metric at seed~42 (Wilcoxon signed-rank: stat\,=\,0.0,
$p$\,=\,0.004), it is far too small to be a headline gain and does \emph{not} translate to the
downstream metric. Measured directly under the held-out nested protocol, the parameterized pool's
\tabpfn{} test accuracy differs from the discrete pool's by only $\Delta{=}{-}0.0002$ across the ten
datasets (paired Wilcoxon $p{=}0.77$, 8 seeds; parameterization improves $4/10$)---no downstream
benefit. The value of parameterization is therefore the structural finding that the optimal KNN $k$ is
dataset-specific, not a task-level accuracy gain.}

\subsection{Transfer Learning}
\label{sec:exp_c6}

\textbf{Hypothesis (C6).} A PPO policy pre-trained on a source dataset and fine-tuned on three
held-out targets reaches within 5\% of a from-scratch policy's reward in $\leq 2{,}000$ steps,
indicating dataset-agnostic transferable structure.

\begin{figure}[!htb]
\centering
\includegraphics[width=\linewidth]{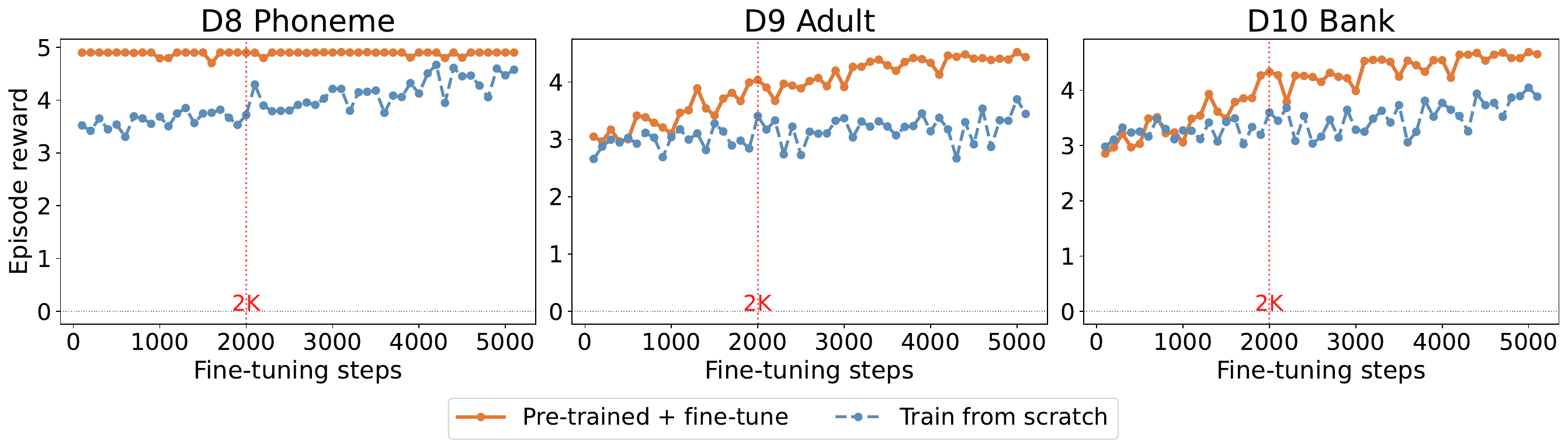}
\caption{Episode reward vs.\ fine-tuning steps on three held-out datasets. Orange: pre-trained on
\texttt{ionosphere} then fine-tuned; blue dashed: from scratch; red dotted: 2K checkpoint \rev{(seed 42)}.}
\label{fig:c6_transfer_curve}
\end{figure}

Figure~\ref{fig:c6_transfer_curve} and Table~\ref{tab:c6_transfer} report episode reward vs.\
fine-tuning steps for the pre-trained and scratch policies on the three held-out datasets. On all
three, the pre-trained policy \emph{exceeds the scratch policy's 5{,}000-step final reward already at
2{,}000 steps}: \texttt{phoneme} $+7.0\%$, \texttt{adult} $+17.2\%$, \texttt{bank-marketing} $+11.5\%$
(rising to $+7.2\%$, $+28.8\%$, $+19.8\%$ after full fine-tuning). The \texttt{adult} gap is notable---$48{,}842$ rows with natural missingness and categoricals---yet the \texttt{ionosphere}-pretrained
policy generalises with no task-specific changes. The consistent pattern indicates the policy
internalises a general prior-alignment strategy (row-preserving imputers, avoiding
distribution-distorting scalers) that \emph{accelerates learning}: fine-tuning reaches a better
solution in $60\%$ fewer steps than scratch.

\input{tab_c6_transfer}

\subsection{\rev{Scalability}}
\label{sec:scalability}
\rev{Because \ours{} queries \tabpfn{} inside the reward, its cost could scale with $n$.
Table~\ref{tab:scalability} fits $t{\propto}n^{a}$ per reward term: the \tabpfn{}-accuracy term is
near-constant ($a{\approx}0$, as \tabpfn{} caps its context at $512$ rows) while the RF, drift and
quality terms grow $\sim$linearly ($a{=}0.88,1.03,1.00$). The model-aware reward is thus
\emph{asymptotically} more scalable than the RF proxy---though not yet in absolute terms: over
$n{\le}20$K the flat \tabpfn{} term (${\approx}20$\,s) still costs $25$--$250\times$ the RF term. The
greedy \tabpfn{} evaluation is embarrassingly parallel and bounded by inner-validation subsampling.}

\begin{table}[t]
\centering
\footnotesize
\caption{\rev{Per-reward-term latency (s) vs.\ $n$ (mean, 3 seeds; power law $t{\propto}n^{a}$,
$R^2{\ge}0.76$). The \tabpfn{} term is flat while the classical terms grow $\sim$linearly, the relative overhead falls ${\sim}240\times{\to}{\sim}4\times$ as $n$ grows.}}
\label{tab:scalability}
\rev{\begin{tabular}{rrrrr}
\toprule
$n$ & \tabpfn{}-acc & RF-acc & drift & \tabpfn{}/RF \\
\midrule
$1{,}000$   & $20.2$ & $0.08$ & $0.02$ & $243\times$ \\
$20{,}000$  & $19.9$ & $0.71$ & $0.31$ & $28\times$ \\
$100{,}000$ & $19.8$ & $4.50$ & $1.79$ & $4.4\times$ \\
\midrule
$a$ & $-0.004$ & $0.88$ & $1.03$ & --- \\
\bottomrule
\end{tabular}}
\end{table}

\section{Discussions}
\label{sec:discussion}

\tabpfn{} applies a fixed internal preprocessing pipeline
(z-normalization, power transform, NaN masks) unconditionally~\cite{hollmann2025tabpfn}, so
model-aware cleaning is \emph{complementary}, and the quadratic retention penalty ($\alpha{=}2$)
reflects the $\mathcal{O}(1/\sqrt{n})$ uncertainty of in-context learners---favouring imputation over
row deletion. Unlike constraint-based systems, \sys{} is model-aware---its reward is driven by the
TFM's own forward-pass accuracy and calibration, with a distance-to-dirty stability term (not the
prior)---and learns a policy that transfers across datasets; because
\ours{} is non-monotone, monotone-pruning arguments do not apply. As limitations, we evaluate
synthetic injection (plus three naturally-dirty sets) on classification only; the greedy oracle scales
as $\mathcal{O}(|\mathcal{A}|^L)$; per-step \tabpfn{} inference (${\sim}0.3$\,s) rules out streaming;
and reward weights are \tabpfn{}-version-specific.
  
\section{Conclusion}
\label{sec:conclusion}

 We presented \sys{}, a deep RL framework for \emph{model-aware} tabular data cleaning for
  Tabular Foundation Models, motivated by the goal of \emph{prior alignment}. Our reward
  taxonomy (C1) shows that na\"ive reward choices are unreliable---three of seven candidates
  collapse to degenerate strategies. Under the holdout protocol, the proposed {\it TFMAwareReward}
  reshapes pipeline selection on all ten datasets (agreeing on only $31\%$ of dataset--seed
  cells) but only \emph{matches} the RF-reward on accuracy; its benefit is confined to
  minority-class macro-$F_1$ under class imbalance, part of which is a reward-agnostic
  calibrated-threshold effect. Calibration is essentially unchanged by the reward
  (C3: 8-seed ECE change $-0.0003$, $p{=}0.83$), and a policy
  pre-trained on one dataset transfers to held-out datasets (C6). Crucially, the implemented
  reward regularizes distance to the \emph{dirty} data, not to the TFM prior, and a diagnostic
  analysis shows the two are positively coupled---the available operators move data \emph{away}
  from the prior. We therefore do not claim to have achieved prior alignment; we frame it as the
  motivating objective and identify the key next step: a reward built on a \emph{faithful}
  approximation of the TFM prior rather than the dirty-data reference, together with adaptation
  to other TFMs by recalibrating the context-size exponent.

\bibliographystyle{ACM-Reference-Format}
\bibliography{refs}

\end{document}

%% file: sec_5_2_c1_rewrite.tex
\subsection{Reward Function Taxonomy}
\label{sec:exp_c1}

\textbf{Hypothesis (C1).}
Among the seven reward functions in \sys{}, drift-penalising and
multi-objective rewards yield higher best-pipeline \tabpfn{} accuracy than
single-metric rewards when evaluated by a fixed greedy search over 112 valid
pipeline sequences on MCAR 15\%-corrupted data.

\begin{figure}[!htb]
\centering
\includegraphics[width=\linewidth]{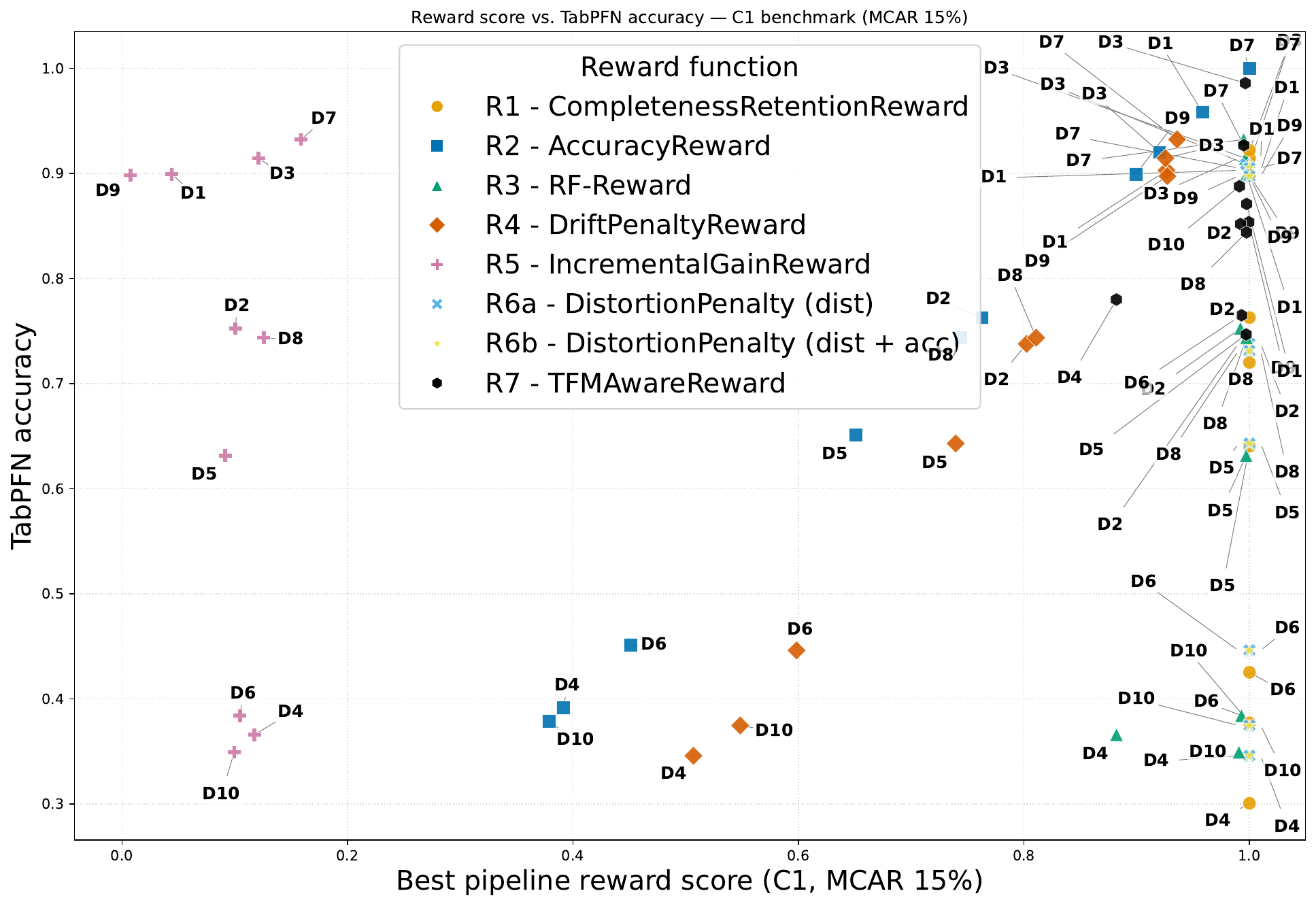}
\caption{Best-pipeline reward score vs.\ \tabpfn{} accuracy per (dataset, reward) pair (MCAR 15\%).
\rev{Collapsed rewards saturate at $x{\approx}1.0$ but span a wide accuracy $y$: R1 $y{\in}[0.30,0.92]$,
R6a/R6b $y{\in}[0.35,0.91]$ (all mean ${\approx}0.69$)---their $y$-coordinate, not $x$, distinguishes the
datasets} \rev{(seed 42)}.}
\label{fig:c1_scatter_per_dataset}
\end{figure}

\noindent\rev{\textbf{Terminology.} We use four
terms precisely. A reward \emph{collapses} if its reward-maximizing pipeline is a
degenerate transformation (e.g.\ delete all rows, or apply no operator) whose
selection is invariant across datasets. A \emph{trivial} reward is one maximized
by a fixed, data-independent pipeline (e.g.\ imputing every cell to saturate a
completeness term), so it carries no discriminative signal across datasets. A
\emph{reward--quality disconnect} occurs when the reward is near its maximum while
the downstream task metrics it is meant to improve---\tabpfn{} test accuracy,
macro-$F_1$, and calibration (ECE)---do not correspondingly improve; equivalently,
gains in the reward do not translate into gains on \emph{any} of these metrics,
independent of their absolute scale. In Figure~\ref{fig:c1_scatter_per_dataset}, we plot accuracy as the
representative metric, but the definition applies to macro-$F_1$ and ECE alike,
which matters for the imbalanced datasets where accuracy and $F_1$ diverge. A
reward is \emph{anti-cleaning} if its optimum prefers less or no cleaning (a
distortion-minimizing no-op), so higher reward corresponds to weaker cleaning.
Collapse and triviality are properties of \emph{which} pipeline is selected; the
disconnect is a property of the reward--metric relationship.}

Figure~\ref{fig:c1_scatter_per_dataset} plots best-pipeline reward score against
\tabpfn{} accuracy for each (dataset, reward) pair; the search is fixed
(exhaustive over 112 ordered sequences), so score differences reflect how each
reward \emph{ranks} pipelines rather than the RL optimizer. A high reward does
\emph{not} imply high accuracy. Three rewards collapse to trivial pipelines that
saturate the reward yet scatter widely on accuracy: R1~(\emph{CompletenessRetentionReward})
scores $1.000{\pm}0.000$ on all ten datasets---any imputation maximises
completeness$\times$retention---and R6a/R6b~(Distortion variants) score
${\approx}1.0$ by selecting \texttt{no-op}, since minimising distortion is exactly
achieved by not cleaning. R4~(\emph{DriftPenaltyReward}) is near-trivial (\texttt{no-op}
on 7/10 datasets): the Wasserstein term dominates the accuracy component, as any
cleaning shifts the distribution away from the reference. R5~(\emph{IncrementalGainReward})
scores an order of magnitude below all others---its step-delta credit assignment
yields vanishing pipeline-level scores in the greedy setting. Only
R2~(\emph{AccuracyReward}, mean $0.72{\pm}0.24$) and R3~(\emph{RF-Reward},
$0.98{\pm}0.04$) are genuinely discriminative; R3 is significantly better than R2
($p{=}0.004$) and R4 ($p{=}0.002$) and selects \texttt{impute(knn)} on 9/10 datasets,
jointly satisfying accuracy, retention and quality.
\rev{Geometrically, the tight high-reward cluster at $x{\approx}1.0$ is therefore \emph{not} itself a
sign of collapse: R3 (and, in C2, R7) sit there with high reward \emph{and} high accuracy---reward
gains that track the metric---whereas R1/R6a/R6b sit there with the reward saturated but accuracy
scattered (collapse and reward--quality disconnect). The terminology above is exactly what separates
the two.}
These findings motivate
\ours{}~(R7): we replace R3's random-forest evaluator with \tabpfn{} to obtain a
reward that measures calibration under the prior-alignment objective
(C2, C3).

%% file: tab_c6_transfer.tex
\begin{table}[!ht]
\centering
\footnotesize
\caption{Reward at the 2K-step checkpoint, fine-tuned from \texttt{ionosphere} vs.\ scratch, on 3
held-out datasets. Gap $=(r_\text{scratch}-r_\text{finetune})/|r_\text{scratch}|$; negative $=$ fine-tune leads \rev{(seed 42)}.}
\label{tab:c6_transfer}
{\begin{tabular}{lcccc}
\toprule
Dataset & Fine-tune  & Scratch  & Gap & Scratch  \\
& @2K &  @2K &  &  final \\
\midrule
\texttt{phoneme}       & \textbf{4.897} & 3.718 & $-31.7\%$ & 4.575 \\
\texttt{adult}         & \textbf{4.037} & 3.408 & $-18.5\%$ & 3.444 \\
\texttt{bank-marketing}    & \textbf{4.332} & 3.601 & $-20.3\%$ & 3.885 \\
\bottomrule
\end{tabular}}
\end{table}